\documentclass{article}

\usepackage{PRIMEarxiv}

\usepackage[utf8]{inputenc} 
\usepackage[T1]{fontenc}    
\usepackage{hyperref}       
\usepackage{url}            
\usepackage{booktabs}       
\usepackage{amsfonts}       
\usepackage{nicefrac}       
\usepackage{microtype}      
\usepackage{lipsum}
\usepackage{fancyhdr}       
\usepackage{graphicx}       
\graphicspath{{media/}}     

\title{Analysis of Knowledge Transfer in  Kernel Regime
\thanks{This work was partially supported by the Wallenberg AI, Autonomous Systems and Software Program (WASP) funded by the Knut and Alice Wallenberg Foundation.} 
}

\author{
   Ashkan Panahi \\
  Chalmers University of Technology\\
  Gothenburg \\
  Sweden\\
  \texttt{ashkan.panahi@chalmers.se} \\
   \And
  Arman Rahbar \\
  Chalmers University of Technology\\
  Gothenburg \\
  Sweden\\
  \texttt{armanr@chalmers.se} \\
   \And
   Chiranjib Bhattacharyya \\
  Indian Institute of Science\\
  Bangalore \\
  India\\
  \texttt{chiru@iisc.ac.in} \\
   \And
   Devdatt Dubhashi \\
  Chalmers University of Technology\\
  Gothenburg \\
  Sweden\\
  \texttt{dubhashi@chalmers.se} \\
   \And
   Morteza Haghir Chehreghani \\
  Chalmers University of Technology\\
  Gothenburg \\
  Sweden\\
  \texttt{morteza.chehreghani@chalmers.se} \\
}
\usepackage{float}
\usepackage{amsthm}
\usepackage{microtype}
\usepackage{graphicx, nicefrac}
\usepackage{subfigure}
\usepackage{booktabs} 
\usepackage{amsmath,bm}
\usepackage{verbatim}
\usepackage{lipsum,enumitem}
\usepackage{graphicx}
\usepackage{babel,blindtext}
\usepackage{balance}
\usepackage{etoolbox}

\newtheorem{theorem}{Theorem}

\newtheorem{lemma}{Lemma}

\usepackage{bm}

\begin{document}
\newcommand{\bGamma}{\bm{\Gamma}}
\newcommand{\bDelta}{\bm{\Delta}}
\newcommand{\bSigma}{\bm{\Sigma}}
\newcommand{\bOmega}{\bm{\Omega}}

\newcommand{\bdelta}{\bm{\delta}}
\newcommand{\bomega}{\bm{\omega}}
\newcommand{\bgamma}{\bm{\gamma}}
\newcommand{\bepsilon}{\bm{\epsilon}}
\newcommand{\blambda}{\bm{\lambda}}
\newcommand{\btheta}{\bm{\theta}}
\newcommand{\bpsi}{\bm{\psi}}
\newcommand{\bmeta}{\bm{\eta}}
\newcommand{\bzeta}{\bm{\zeta}}
\newcommand{\bmu}{\bm{\mu}}
\newcommand{\bnu}{\bm{\nu}}
\newcommand{\bpi}{\bm{\pi}}
\newcommand{\bsigma}{\bm{\sigma}}
\newcommand{\bphi}{\bm{\phi}}

\newcommand{\barbphi}{\bar{\bphi}}

\newcommand{\tilDelta}{\tilde{\Delta}}
\newcommand{\tlDelta}{\tilde{\Delta}}
\newcommand{\tlepsilon}{\tilde{\epsilon}}
\newcommand{\tltheta}{\tilde{\theta}}
\newcommand{\tlphi}{\tilde{\phi}}

\newcommand{\bA}{\mathbf{A}}
\newcommand{\bC}{\mathbf{C}}
\newcommand{\bD}{\mathbf{D}}
\newcommand{\bE}{\mathbf{E}}
\newcommand{\bF}{\mathbf{F}}
\newcommand{\bG}{\mathbf{G}}
\newcommand{\bH}{\mathbf{H}}
\newcommand{\bI}{\mathbf{I}}
\newcommand{\bJ}{\mathbf{J}}
\newcommand{\bL}{\mathbf{L}}
\newcommand{\bM}{\mathbf{M}}
\newcommand{\bN}{\mathbf{N}}
\newcommand{\bP}{\mathbf{P}}
\newcommand{\bQ}{\mathbf{Q}}
\newcommand{\bR}{\mathbf{R}}
\newcommand{\bS}{\mathbf{S}}
\newcommand{\bT}{\mathbf{T}}
\newcommand{\bU}{\mathbf{U}}
\newcommand{\bV}{\mathbf{V}}
\newcommand{\bW}{\mathbf{W}}
\newcommand{\bX}{\mathbf{X}}
\newcommand{\bY}{\mathbf{Y}}
\newcommand{\bZ}{\mathbf{Z}}

\newcommand{\ba}{\mathbf{a}}
\newcommand{\bb}{\mathbf{b}}
\newcommand{\bd}{\mathbf{d}}
\newcommand{\be}{\mathbf{e}}
\newcommand{\mbf}{\mathbf{f}}
\newcommand{\bg}{\mathbf{g}}
\newcommand{\bh}{\mathbf{h}}
\newcommand{\bl}{\mathbf{l}}
\newcommand{\bn}{\mathbf{n}}
\newcommand{\bp}{\mathbf{p}}
\newcommand{\bq}{\mathbf{q}}
\newcommand{\br}{\mathbf{r}}
\newcommand{\bs}{\mathbf{s}}
\newcommand{\bu}{\mathbf{u}}
\newcommand{\bv}{\mathbf{v}}
\newcommand{\bw}{\mathbf{w}}
\newcommand{\bx}{\mathbf{x}}
\newcommand{\by}{\mathbf{y}}
\newcommand{\bz}{\mathbf{z}}

\newcommand{\rmm}{\mathrm{m}}

\newcommand{\hbeta}{\hat{\beta}}
\newcommand{\htheta}{\hat{\theta}}
\newcommand{\hsigma}{\hat{\sigma}}

\newcommand{\hb}{\hat{b}}
\newcommand{\hp}{\hat{p}}
\newcommand{\hr}{\hat{r}}
\newcommand{\hs}{\hat{s}}
\newcommand{\hw}{\hat{w}}
\newcommand{\hx}{\hat{x}}

\newcommand{\hN}{\hat{N}}

\newcommand{\hbSigma}{\hat{\bm{\Sigma}}}

\newcommand{\hba}{\hat{\mathbf{a}}}
\newcommand{\hbs}{\hat{\mathbf{s}}}
\newcommand{\hbx}{\hat{\mathbf{x}}}
\newcommand{\hbv}{\hat{\mathbf{v}}}

\newcommand{\hbW}{\hat{\mathbf{W}}}

\newcommand{\dif}{\text{d}}

\newcommand{\bbC}{\mathbb{C}}
\newcommand{\bbE}{\mathbb{E}}
\newcommand{\bbR}{\mathbb{R}}
\newcommand{\bbN}{\mathbb{N}}
\newcommand{\bbZ}{\mathbb{Z}}

\newcommand{\calA}{\mathcal{A}}
\newcommand{\calB}{\mathcal{B}}
\newcommand{\calC}{\mathcal{C}}
\newcommand{\calD}{\mathcal{D}}
\newcommand{\calE}{\mathcal{E}}
\newcommand{\calF}{\mathcal{F}}
\newcommand{\calH}{\mathcal{H}}
\newcommand{\calL}{\mathcal{L}}
\newcommand{\calN}{\mathcal{N}}
\newcommand{\calM}{\mathcal{M}}
\newcommand{\calP}{\mathcal{P}}
\newcommand{\calS}{\mathcal{S}}
\newcommand{\calT}{\mathcal{T}}
\newcommand{\calV}{\mathcal{V}}
\newcommand{\calW}{\mathcal{W}}
\newcommand{\calX}{\mathcal{X}}
\newcommand{\calY}{\mathcal{Y}}

\newcommand{\calhL}{\mathcal{\hat{L}}}

\newcommand{\tlA}{\tilde{A}}
\newcommand{\tlC}{\tilde{C}}
\newcommand{\tlD}{\tilde{D}}

\newcommand{\tlv}{\tilde{v}}
\newcommand{\tls}{\tilde{s}}
\newcommand{\tlx}{\tilde{x}}

\newcommand{\bara}{\bar{a}}
\newcommand{\barb}{\bar{b}}
\newcommand{\barg}{\bar{g}}
\newcommand{\barm}{\bar{m}}
\newcommand{\barn}{\bar{n}}
\newcommand{\barr}{\bar{r}}
\newcommand{\bary}{\bar{y}}

\newcommand{\barC}{\bar{C}}
\newcommand{\barD}{\bar{D}}
\newcommand{\barH}{\bar{H}}
\newcommand{\barK}{\bar{K}}
\newcommand{\barL}{\bar{L}}
\newcommand{\barW}{\bar{W}}

\newcommand{\barba}{\bar{\ba}}
\newcommand{\barbg}{\bar{\bg}}
\newcommand{\barbw}{\bar{\bw}}
\newcommand{\barbx}{\bar{\bx}}
\newcommand{\barby}{\bar{\by}}
\newcommand{\barbz}{\bar{\bz}}

\newcommand{\barbH}{\bar{\bH}}

\newcommand{\tlbA}{\tilde{\bA}}
\newcommand{\tlbE}{\tilde{\bE}}
\newcommand{\tlbH}{\tilde{\bH}}
\newcommand{\tlbW}{\tilde{\bW}}

\newcommand{\tlbf}{\tilde{\mbf}}
\newcommand{\tlbv}{\tilde{\bv}}
\newcommand{\tlbw}{\tilde{\bw}}

\newcommand{\tc}{\text{c}}
\newcommand{\td}{{\text{d}}}

\newcommand{\bzero}{\mathbf{0}}
\newcommand{\bone}{\mathbf{1}}

\newcommand{\suml}{\sum\limits}
\newcommand{\minl}{\min\limits}
\newcommand{\maxl}{\max\limits}
\newcommand{\infl}{\inf\limits}
\newcommand{\supl}{\sup\limits}
\newcommand{\liml}{\lim\limits}
\newcommand{\intl}{\int\limits}
\newcommand{\bigcupl}{\bigcup\limits}

\newcommand{\opconv}{\text{conv}}

\newcommand{\eref}[1]{(\ref{#1})}

\newcommand{\sinc}{\text{sinc}}
\newcommand{\tr}{\text{Tr}}
\newcommand{\var}{\text{Var}}
\newcommand{\cov}{\text{Cov}}
\newcommand{\tth}{\text{th}}

\newcommand{\nwl}{\nonumber\\}

\newcommand{\prox}{\mathrm{prox}}

\newenvironment{vect}{\left[\begin{array}{c}}{\end{array}\right]}

\maketitle
\begin{abstract}
  Knowledge transfer is shown to be a very successful technique for training neural classifiers: together with the ground truth data, it uses the "privileged information" (PI) obtained by a "teacher" network to train a "student" network. It has been observed that classifiers learn much faster and more reliably via knowledge transfer. However, there has been little or no theoretical analysis of this phenomenon. To bridge this gap, we propose to approach the problem of knowledge transfer by regularizing the fit between the teacher and the student with PI provided by the teacher. Using tools from dynamical systems theory, we show that when the student is an extremely wide two layer network, we can analyze it in the kernel regime and show that it is able to interpolate between PI and the given data. This characterization sheds new light on the relation between the training error and capacity of the student relative to the teacher. Another contribution of the paper is a quantitative statement on the convergence of student network. We prove that the teacher reduces the number of required iterations for a student to learn, and consequently improves the generalization power of the student.  We give corresponding experimental analysis that validates the theoretical results and yield additional insights.
\end{abstract}




\section{Introduction}
Knowledge transfer considers improving learning processes by leveraging the knowledge learned  from other tasks or trained models. Several studies have demonstrated the  effectiveness of knowledge transfer in different settings. For instance, in \cite{chen2017learning}, knowledge transfer has been used to improve object detection models. 
Knowledge transfer has been applied in different levels to neural machine translation in \cite{KD-sequence}. It has also been employed for Reinforcement learning in \cite{rlNips}. Recommender systems can also benefit from knowledge transfer as shown in \cite{auto-KD}.

An interesting case of knowledge transfer is \emph{privileged information} \cite{VapnikV09,VapnikI17} where the goal is to supply a student learner with privileged information during training session.
Another special case is concerned with knowledge distillation \cite{hinton15} which suggests to train classifiers using the real--valued outputs of another classifier as target values than using actual ground--truth labels.
These two paradigms are unified within a consistent framework in \cite{LopezPaz15}. 

The privileged information paradigm introduced in \cite{VapnikV09} aims at mimicking some elements of human teaching in order to improve the process of learning with examples. In particular, the teacher provides some additional information for the  learning task along with training examples. This privileged information is only available during the training phase. The work in \cite{privilegeNips} outlines the theoretical conditions required for the additional information from a teacher to a student. If a teacher satisfies these conditions, it will help to accelerate the learning rate. Two different mechanisms of using privileged information are introduced in \cite{JMLRPrivileged}. In the first mechanism, the concept of similarity in the training examples in the student is controlled. In the second one, the knowledge in the space of privileged information is \emph{transferred} to the space where the student is working.

In 2014, Hinton et al. \cite{hinton15} studied the effectiveness of knowledge distillation. They
showed that it is easier to train classifiers using the real--valued outputs of another classifier as target values than using actual ground--truth labels. They introduced the term \emph{knowledge distillation} for this phenomenon. Since then, distillation--based training has been confirmed in several different types of neural networks \cite{chen2017learning,yim2017gift,yu2017visual}. It has been observed that 
optimization is generally more well--behaved than
with label-based training, and it needs
less regularization or specific optimization tricks.

While the practical benefits of knowledge transfer in neural networks (e.g. via distillation) are beyond doubt,
its theoretical justification remains almost completely unclear.
Recently, Phuong and Lampert \cite{phuong19} made an attempt to analyze knowledge distillation in a simple model. In their setting, both the teacher and the student are \emph{linear} classifiers (although the student's weight vector is allowed a over-parametrized representation as a product of matrices). They give conditions under which the student's weight vector converges (approximately) to that of the teacher and derive consequences for generalization error. Crucially, their analysis is limited to linear networks. 

Knowledge transfer in neural networks and privileged information are related through a unified framework proposed in  \cite{LopezPaz15}. In this work knowledge transfer in neural networks is examined from a theoretical perspective via casting that as a form of learning with privileged information. However, \cite{LopezPaz15} uses a heuristic argument for the effectiveness of knowledge transfer with respect to generalization error rather than a rigorous analysis.

In our knowledge transfer analysis, we assume the so-called \emph{kernel regime}. A series of recent works,  e.g., \cite{Arora19,DuH19,CaoG19a,mei2019mean} achieved breakthroughs in understanding how (infinitely) wide neural network training behaves in this regime, where the dynamics of training by gradient descent can be approximated by the dynamics of a linear system. We extend the repertoire of the methods that can be applied in such settings.

{\bf Contributions}: We carry out a theoretical analysis of knowledge transfer which consistently covers aspects of privileged information for non-linear neural networks. We situate ourselves in the recent line of work that analyzes the dynamics of neural networks under the kernel regime. It was shown that the behaviour of training by gradient descent (GD) in the limit of very wide neural networks can be  approximated by \emph{linear system dynamics}. This is dubbed the \emph{kernel regime} because it was shown in \cite{JFH18} that a fixed kernel -- the \emph{neural tangent kernel} -- characterizes the behavior of fully-connected infinite width neural networks in this regime.

Our framework is general enough to encompass Vapnik's notion of \emph{privileged information} and provides a unified analysis of \emph{generalized distillation} in the paradigm of \emph{machines teaching machines} as in \cite{LopezPaz15}. For this analysis, we exploit new tools that go  beyond the previous techniques in the literature, as in \cite{Arora19,DuH19,CaoG19a,mei2019mean} - we believe these new tools will contribute to further development of the nascent theory. Our main results are:\\ 
i) We formulate the knowledge transfer problem as a least squares optimization problem with regularization provided by privileged knowledge.
This allows us to  characterize precisely what is learnt by the student network in Theorem~\ref{thm:1} showing that the student converges to an interpolation between the data and the privileged information guided by the strength of the regularizer.\\ 
ii) We characterize the speed of convergence in Theorem~\ref{thm:3} in terms of the overlap between a combination of the label vector and the knowledge vectors and the spectral structure of the data as reflected by the vectors. \\
iii) We introduce novel techniques from systems theory, in particular, Laplace transforms of time signals to analyze time dynamics of neural networks. The recent line of work e.g. in \cite{Arora19,DuH19,CaoG19a,mei2019mean} has highlighted the dynamical systems view in statistical learning by neural networks. We introduce a more coherent framework with a wider ranger of techniques to exploit this view point. This is in fact necessary since the existing approaches are insufficient in our case because of the asymmetry in the associated Gram matrix and its complex eigen-structure. We use the poles of the Laplace transform to analyze the dynamics of the training process in section~\ref{sec:dynamics}.  \\
iv)
We discuss the relation of the speed of convergence to the generalization power of the student and show that the teacher may improve the generalization power of the student by speeding up its convergence, hence effectively reducing its capacity.
\\
v) We experimentally demonstrate different aspects of our  knowledge transfer framework supported by our theoretical analysis. 
We exploit the data overlap characterization in Theorem~\ref{thm:3} using optimal kernel-target alignment \cite{CortesMohri2012} to compute kernel embeddings which lead to better knowledge transfer. 
\vspace{-1mm}
\section{Problem Formulation and Main Results}
We study knowledge transfer in an analytically tractable setting: the two layer non--linear model studied in \cite{Arora19,du2018gradient,DuH19,CaoG19a}:
\begin{equation}\label{eq:network}
    f(\bx)=\suml_{k=1}^m\frac{a_k}{\sqrt m}\sigma(\bw_k^T\bx),
\end{equation}
Here the weights $\{\bw_k\}_{k=1}^m$ are the model variables  corresponding to $m$ hidden units, $\sigma(\ldotp)$ is a real (nonlinear) activation function and the weights $\{a_k\}$ are fixed. While we assume that the student network maintains the form in \eqref{eq:network} throughout this paper, the teacher may not assume a definite architecture. However, we often specialize our results for the case that the teacher also takes the form in \eqref{eq:network} with a larger number $\barm$ of hidden units. 

 For training the student, we introduce a general optimization framework that considers knowledge transfer. Given a dataset $\{(\bx_i,y_i)\}_{i=1}^n$ comprising of $n$ data samples $\{\bx_i\}$ and their corresponding labels $\{y_i\}$, our framework is given by
 \begin{equation}\label{eq:KT_main}
    \minl_{\{\bw_k\}}\suml_i(y_i-f(\bx_i))^2+\lambda\suml_i\suml_k\left(\phi^{(k)}(\bx_i)-f^{(k)}(\bx_i)\right)^2,
\end{equation}
where $f(\ldotp)$ is stated in \eqref{eq:network} and $f^{(k)}(\bx)=\sigma(\bw_k^T\bx)$ is the corresponding $k^\tth$ hidden feature of the student network. As seen in \eqref{eq:KT_main}, our framework consists of a least squares optimization problem: 
$\min\sum(y_i-f(\bx_i))^2$
with an additional regularization term incorporating the teacher's knowledge represented by the \emph{privileged knowledge} terms $\phi^{(k)}$. The coefficient $\lambda\geq 0$ is the regularization parameter.

Our analysis considers generic forms of the privileged knowledge functions $\phi^{(k)}$. 
However, we are particularly interested in a setup where these functions are selected from the hidden neurons of a pre-trained teacher. More precisely, $\phi^{(k)}(\bx)=\sigma(\langle\bw_k^{\mathrm{teacher}},\bx\rangle)$ where $\bw_k^{\mathrm{teacher}}$ is the trained weight of the $k^\tth$ selected unit of the teacher with the architecture in \eqref{eq:network}. Note that $\left\{\bw_k^{\mathrm{teacher}}\right\}$  is a subset of the teacher weights. This case is closely connected to a well-known knowledge distillation (KD) setup 
 empirically studied in e.g. \cite{chen2017learning}.

 We study the generic behavior of the gradient descent (GD) algorithm when applied to the optimization in \eqref{eq:KT_main}. In the spirit of the analysis in \cite{du2018gradient,Arora19}, shortly explained in Section \ref{sec:previous_work}, we carry out an investigation on the dynamics for GD that answers two fundamental  questions:
\emph{i. What does the (student) network learn?} 
\emph{ii. How fast is the convergence by the gradient descent?}
The answer to both these questions emerges from the analysis of the dynamics of GD. From the perspective of KD, this approach complements related recent studies, such as \cite{phuong19} that address similar questions. However, our work is different, as \cite{phuong19} is limited to a single hidden unit ($m=1$), Sigmoid activation $\sigma$ and cross-entropy replacing the square-error loss. While convexity plays a major role in \cite{phuong19}, our analysis concerns the non-convex setup in \eqref{eq:KT_main} with further conditions on initialization. Additionally, our result is applicable to a different regime with a large number $m$ of units and high expression capacity.

%

\section{Main Results on Dynamics}
\subsection{General Linear Systems Theory Framework}
\label{sec:general_framework}
The existing analysis of dynamics for neural networks in a series of recent papers \cite{Arora19,DuH19,CaoG19a} is tied centrally to the premise that the behaviour of GD for the optimization can be approximated by a linear dynamics of finite order.
To isolate the negligible effect of learning rate $\mu$ in GD, it is also conventional to study the case $\mu\to 0$ where GD is alternatively represented by an ordinary differential equation (ODE), known as the gradient flow, with a continuous "time" variable $t$ replacing the iteration number $r$ (being equivalent to the limit of $r\mu$). Let us denote by $\mbf(t)$ the vector of the output $f(\bx_i,t)$ of the network at time $t$. Then, the theory of linear systems with a finite order suggests the following expression for the evolution of $\mbf(t)$:
\begin{equation}
    \mbf(t)=\mbf_\infty+\bdelta(t),
\end{equation}
where $\mbf_\infty$ is a constant and
\begin{equation}\label{eq:dynamics_template}
    \bdelta(t)=\bu_1e^{-p_1t}+\bu_2e^{-p_2t}+\ldots+\bu_de^{-p_dt}.
\end{equation}
Here, $d$ is the order of the linear system and complex-valued vectors $\bu_1,\ldots,\bu_m$ and nonzero complex values $p_1,p_2,\ldots,p_d$ are to be determined by the specifications of the dynamics. The constants $\{p_j\neq 0\}$ are called poles, that also correspond to the singular points of the Laplace transform $\bF(s)$ of $\mbf(t)$ (except for $0$, which corresponds to the constant $\mbf_\infty$ in our formulation). We observe that such a representation may only have a convergence (final) value at $t\to\infty$ if the poles have strictly positive real parts, in which case $\mbf_\infty$ is the final value. Moreover, the asymptotic rate of convergence is determined by the dominating term in \eqref{eq:dynamics_template}, i.e. the  smallest value $\Re(p_j)$ with a nonzero vector $\bu_j$. We observe that identifying $\mbf_\infty$ and the dominating term responds to the aforementioned questions of interest. In this paper, we show that these values can be calculated as the number $m$ of hidden units increases. 

{\bf Definitions}: Let us take $\bw_k=\bw_k(0)$ as the initial values of the weights and define $\bH_k=(\sigma^\prime(\bw_k^T\bx_i)\sigma^\prime(\bw_k^T\bx_i)\bx_i^T\bx_j)$ as the $k^\tth$ realization of the "associated gram matrix" where $\sigma^\prime$ denotes the derivative function of $\sigma$ (that can be defined in the distribution sense). Further, denote by $\mbf^k(0)$ the vector of the initial values $f^k(\bx_i)=\sigma(\bw_k^T\bx_i)$ of the $k^\tth$ unit for different data points $\{\bx_i\}$ and take $a=\sum_k\nicefrac{a_k^2}{m}$. Finally, take $p_1,p_2,\ldots,p_d$ for $d=n\times m$ as positive values where at $s=-p_i$, the value $-1$ is the eigenvalue of the matrix 
$\bT(s)=\suml_k\frac{a_k^2}m(s\bI+\lambda\bH_k)^{-1}\bH_k, $
 with $\bv^1,\bv^2,\ldots,\bv^d$ being the corresponding eigenvectors ($\bT(s)$ is symmetric). 
\subsection{What does the student learn?}
\label{sec:learnt}
This result pertains to the first question above,  concerning the final value of $\mbf$. For this, we prove the following result:
\begin{theorem}\label{thm:1}
Suppose that $m$ is large and $\|\bphi_k-\mbf_k(0)\|=O(1/m)$. Under mild conditions (Section \ref{sec:proofs}), it holds that $\liml_{t\to\infty}\mbf(t)=\mbf_\infty$ where
\begin{equation}
    \mbf_\infty=\frac{1}{a+\lambda}\left(a\by+\lambda\suml_k\frac{a_k\bphi_k}{\sqrt{m}}\right)+o_m(1).
\end{equation}
\end{theorem}
This is an intuitive result: the final output of the student is mixture of the true labels $\by$ and the teacher's provided knowledge vectors $\bphi_k$.

{\bf Random Privileged Knowledge Setup:} Indeed, an interesting case is when the teacher is itself a strong predictor of the labels 
    $\bb:=\suml_k\frac{a_k\bphi_k}{\sqrt{m}}=\by+o_m(1)$,
which further yields a perfect prediction by the student. This can be the case when the teacher is a wider network in the form of \eqref{eq:network} with $\barm$ hidden neurons and their corresponding weights $\barbw_l^{\mathrm{teacher}}$ and bounded coefficients $\bara^\mathrm{teacher}_l$ in the second layer. The results in \cite{du2018gradient} guarantee that such a network can be perfectly trainable over the samples. Let us consider the case where $\{a^{\mathrm{teacher}}_k, \bw^{\mathrm{teacher}}_k\}\subset \{\bara^{\mathrm{teacher}}_l,\barbw^{\mathrm{teacher}}_l\}$ is independently randomly selected. Then, we may invoke Theorem \ref{thm:1} by taking
$\phi^{(k)}(\bx)=\sigma(\langle\bw_k^{\mathrm{teacher}},\bx\rangle)$ and $a_k= q a^\mathrm{teacher}_k$, where $q=\sqrt{\frac {\barm}m}$. Then, we have
$    \bbE \left[\bb\right]=\mbf^{\mathrm{teacher}},\  \var[\bb]=O\left(\frac{\barm}{m}-1\right)$. This shows that for large $\barm$ taking a sufficiently large fraction of the units will introduce negligible harm to the student's solution. 

\subsection{How fast does the student learn?}
\label{sec:speed}
Now, we turn our attention to the question of the speed of convergence, for which we have the following result:
\begin{theorem}\label{thm:3}
Define
\[
\mbf^{(k)}_\infty=\frac {a_k}{\lambda\sqrt{m}}(\by-\mbf_\infty)+\bphi_k.
\]
With similar assumptions to Theorem \ref{thm:1} (given in Section \ref{sec:proofs}), the dynamics of $\mbf$ can be written as\footnote{The rate analysis is in $L_1$ sense, i.e. $a(t)=b(t)+o_m(1)$ means $\intl_0^\infty\|a(t)-b(t)\|_2\td t=o_m(1)$.
}
\[
\mbf(t)=\mbf_\infty+\suml_{j=1}^ne^{-p_jt}\alpha_j\bu^{j}+o_m(1),
\quad
\]
where
\[
\alpha_j=\suml_k\frac{a^2_k}{m}\left\langle\bv^j,\bH_k(p_j\bI-\lambda\bH_k)^{-1}(\mbf^{(k)}_\infty-\mbf^{(k)}(0))\right\rangle.
\]
\end{theorem}
A straightforward consequence of Theorem \ref{thm:3} is that 
\[
\|\mbf(t)-\mbf_\infty\|_2=O\left(e^{-p_{\text{min}}t}\right),
\]
where $p_{\text{min}}$ is the minimum value of $p_j$s. In other words, convergence is linear. In practice, the discrete time process of gradient descent with a step size $\mu$ is used. Although our analysis is instead based on the common choice of gradient flow, we remark that with a similar approach, one can show that for a sufficiently small step size, e.g. $\mu<\frac 1{2p_{\text{max}}}$ with $p_{\text{max}}$ being the largest of $p_j$s, the convergence rate remains linear:
\[
\|\mbf_n-\mbf_\infty\|_2=O\left((1-\mu p_{\text{min}})^n\right),
\]
where with a slight abuse of notation, $\mbf_n$ denotes the network output in the $n^\tth$ iteration of GD. 

From the definition, $\alpha_j$ can be interpreted as an average overlap between a combination of the label vector $\by$ and the knowledge vectors $\bphi_k$, and the "spectral" structure of the data as reflected by the vectors $(p_j\bI-\lambda\bH_k)^{-1}\bH_k\bv^j$. This is a generalization of the geometric argument in \cite{Arora19} in par with the "data geometry" concept introduced in \cite{phuong19}. We will later use this result in our experiments to improve knowledge transfer by modifying the data geometry of $\alpha_j$ coefficients.

\subsection{Further Remarks}\label{sec:further}
The above two results have a number of implications on the knowledge transfer process: 

{\bf Extreme Cases:} First, note that the case $\lambda=0$ reproduces the results in \cite{du2018gradient}: The final value simply becomes $\by$ while the poles will become the singular values of the matrix $\bH=\suml_k\frac{a_k^2}{m}\bH_k$. The other extreme case of $\lambda=\infty$ corresponds to pure transfer from the teacher, where the effect of the first term in \eqref{eq:KT_main} becomes negligible and hence the optimization boils down to individually training each hidden unit by $\phi_k$. One may then expect the solution of this case to be $f_k=\phi_k$. However, the conditions of the above theorems become difficult to verify, but we shortly present experiments that numerically investigate the corresponding dynamics.

{\bf Speed-accuracy trade-off:}  As previously pointed out, for a finite value of $\lambda$, the final value $\mbf_\infty$ is a weighted average, depending on the quality of $\bphi^k$. Defining $e=\|\mbf_\infty-\by\|$ as the final error, we simply conclude that $e=\frac\lambda{a+\lambda}\|\suml_k\frac{a_k\bphi_k}m-\by\|$, where the term $\|\suml_k\frac{a_k\bphi_k}m-\by\|$ reflects the quality of teacher in representing the labels. Also for an imperfect teacher the error $e$ monotonically increases with $\lambda$. At the same time, we observe that larger $\lambda$ has a positive effect on the speed of learning.

\begin{theorem}
Given, the above definition, it holds that $p_j\geq \lambda\sigma_0$, where $\sigma_0$ is the smallest nonzero eigenvalue of all $\bH_k$s.
\end{theorem} 

This sets an intuitive trade-off, where increasing $\lambda$, i.e. relying more on the teacher, improves the speed of learning, while magnifying potential teacher's imperfections. Note that $\sigma_0$ remains finite and of $O(\sqrt n)$, even if the number $m$ of $\bH_k$s increases.

{\bf Effect of data geometry:} As we demonstrate in Section \ref{sec:proofs}, the vectors $\bH_k(p_j-\lambda\bH_k)^{-1}\bv_j$ constitute an eigen-basis structure corresoonding to the poles $p_j$. If the data has a small overlap with the eigen-basis  corresponding to small values of $p_j$, then the values $\alpha_j$ for small poles $p_j$ will drop and the dynamics is mainly identified by the large poles $p_j$, speeding up the convergence properties. This defines a notion of a suitable geometry for knowledge transfer.

{\bf On initializing the student:} Finally, the assumption $\|\bphi_k-\mbf^{(k)}(0)\|=O(1/m)$ can be simply satisfied 
with single hidden layer, where we have $\phi_k(\bx)=\sigma(\langle\bw^{\mathrm{teacher}}_k,\bx\rangle)$ and initializing the weights of the student by that of the teacher $\bw_k(0)=\bw^{\mathrm{teacher}}_k$ leads to $\bphi_k=\mbf^{(k)}(0)$. We further numerically investigate the consequences of violating this assumption. 


\subsection{Consequences on Generalization}

In \cite{Arora19}, a bound on the generalization of the NN in \eqref{eq:network}, when trained by GD, is given. Their approach is to show that for GD, the trained weights $\bw_k$ and their initial values satisfy 
\begin{equation}\label{eq:ar}
    \suml_k\|\bw_k-\bw_k(0)\|^2\leq\by^T\bH^{-1}\by+o(1)
\end{equation}
They proceed by showing that the family of such neural notworks have a bounded Rademacher complexity and hence the generalization power. The bound in \eqref{eq:ar} is a natural consequence of the fact that the learning rate of each weight, at each time, is on average proportional to the convergence rate of the network, which is shown to be exponential. In other words, the weights will not have enough time to escape the ball defined by \eqref{eq:ar}.  Our analysis of convergence in Theorem~\ref{thm:3} leads to a similar generalizaton bound for the student network. In fact, as $\lambda$ increases we get faster convergence by Theorem~\ref{thm:3} since the weights have less time to update, leading to a tighter bound.  This claim is intuitive too: as learning relies more on the teacher, less variation is expected, leading to better generalization. 
This provides additional insights on the success of knowledge transfer in practice.

\section{Analysis and Insights}
\label{sec:previous_work}

The study in \cite{du2018gradient} on the dynamics of backpropagation serves as our main source of inspiration, which we review first.  The point of departure in this work is to represent the dynamics of BP or gradient descent (GD) for the standard $\ell_2$ risk minimization, as in \eqref{eq:KT_main} and \eqref{eq:network} with $\lambda=0$.  In this case, the associated ODE to GD reads:
\begin{equation}\label{eq:weight_dynamics}
    \frac{\td\bw_k}{\td t}(t)=\frac{a_k}{\sqrt m}\bL(\bw_k(t))(\by-\mbf(t)),
\end{equation}
where $\by,\mbf(t)$ are respectively the vectors of $\{y_k\}$ and $f(\bx_k)$, calculated in \eqref{eq:network} by replacing $\bw_k=\bw_k(t)$. Moreover, the matrix $\bL(\bw)$ consists of $\sigma^\prime(\bw^T\bx_k)\bx_k$ as its $k^\tth$ column. While the dynamics in \eqref{eq:weight_dynamics} is generally difficult to analyze, we identify two simplifying ingredients  in the study of \cite{du2018gradient}. First, it turns the attention from the dynamics of weights to the dynamics of the function, as reflected by the following relation:
\begin{equation}
    \frac{\td\mbf}{\td t}(t)=\suml_k\frac{a_k}{\sqrt m}\bL^T_k\frac{\td\bw_k}{\td t}=\bH(t)(\by-\mbf(t)),
\end{equation}
where $\bL_k=\bL_k(t)$ is a short-hand notation for $\bL(\bw_k(t))$ and $\bH(t)=\suml_ka_k^2\bL_k^T\bL_k/m$. The second element in the proof can be formulated as follows:

{\bf Kernel Hypothesis (KH):}
In the asymptotic case of $m\to\infty$, the dynamics of $\bH(t)$ has a negligible effect, such that it may be replaced by $\bH(0)$, resulting to a linear dynamics.

The reason for our terminology of the KH is that under this assumption, the dynamics of BP resembles that of a kernel regularized least squares problem. The investigation in \cite{du2018gradient} further establishes KH under mild assumptions and further notes that for random initialization of weights $\bH(0)$ concentrates on its mean value, denoted by $\bH^\infty$. 

\subsection{Dynamics of Knowledge Transfer}\label{sec:dyn}
Following the methodology of \cite{du2018gradient}, we proceed by providing the dynamics of the GD algorithm for the optimization problem in \eqref{eq:KT_main} with $\lambda>0$. Direct calculation of the gradient leads us to the following associated ODE for GD:
\begin{equation}
    \frac{\td \bw_k}{\td t}=\bL_k\left[\frac{a_k}{\sqrt m}(\by-\mbf(t))+\lambda(\bphi^{(k)}-\mbf^{(k)})\right],
\end{equation}
where $\bL_k,\by,\mbf(t)$ are similar to the previous case in \eqref{eq:weight_dynamics}. Furthermore, $\mbf^{(k)},\bphi^{(k)}$ are respectively the vectors of $\{f^{(k)}(\bx_i)\}_i$ and $\{\phi^{(k)}(\bx_i)\}_i$. We may now apply the methodology of \cite{du2018gradient} to obtain the dynamics of the features. We also observe that unlike this work, the hidden features $\{\mbf^{(k)}\}$ explicitly appear in the dynamics:

\begin{eqnarray}\label{eq:exact_dynamics}
    &\frac{\td\mbf^{(k)}}{\td t}(t)=\bL_k^T\frac{\td \bw_k}{\td t}=\nwl 
    &\bH_k(t)\left[\frac{a_k}{\sqrt m}(\by-\mbf(t))+\lambda(\bphi^{(k)}-\mbf^{(k)}(t))\right],
\end{eqnarray}
where $\bH_k(t)=\bL_k^T\bL_k$ and
\begin{equation}\label{eq:output}
    \mbf(t)=\suml_k\frac{a_k}{\sqrt m}\mbf^{(k)}(t).
\end{equation}
This relation will be central in our analysis and we may slightly simplify it by introducing $\bdelta^{(k)}=\mbf^{(k)}-\mbf^{(k)}_\infty$ and $\bdelta=\mbf-\mbf_\infty$.
In this case the dynamics in \eqref{eq:exact_dynamics} and \eqref{eq:output} simplifies to: 

\begin{eqnarray}\label{eq:exact_dynamics_delta}
    &\frac{\td\bdelta^{(k)}}{\td t}(t)=-\bH_k(t)\left[\frac{a_k}{\sqrt m}\bdelta(t)+\lambda\bdelta^{(k)}(t)\right],\nwl
    &\bdelta(t)=\suml_k\frac{a_k}{\sqrt m}\bdelta^{(k)}(t).
\end{eqnarray}
Finally, we give a more abstract view on the relation in \eqref{eq:exact_dynamics_delta} by introducing the block vector $\bmeta(t)$ where the $k^\tth$ block is given by $\bdelta^{(k)}(t)$. Then, we may write \eqref{eq:exact_dynamics_delta} as
$\nicefrac{\td\bmeta}{\td t}(t)=-\barbH(t)\bmeta(t),$
where $\barbH(t)$ is a block matrix with $\bH_k(t)\left(\frac{a_ka_l}m+\lambda\delta_{k,l}\right)$ as its $k,l$ block ($\delta_{k,l}$ denotes the Kronecker delta function).

\subsection{Dynamics Under Kernel Hypothesis: Analysis by Laplace Transform}
\label{sec:dynamics}
Now, we follow \cite{du2018gradient} by simplifying the relation in \eqref{eq:exact_dynamics} under the kernel hypothesis, which in this case assumes the matrices $\barbH(t)$ to be fixed to its initial value $\barbH=\barbH(0)$, leading again to a linear dynamics:
\begin{equation}\label{eq:abstract_dynamics}
\bmeta(t)=e^{-\barbH t}\bmeta(0).
\end{equation}
Despite similarities with the case in \cite{du2018gradient,Arora19}, the relation in \eqref{eq:abstract_dynamics} is not simple to analyze due to the asymmetry in $\barbH$ and the complexity of its eigen-structure. For this reason, we proceed by taking the Laplace transform of \eqref{eq:exact_dynamics_delta} (assuming $\bH_k=\bH_k(t)=\bH_k(0)$) which after straightforward manipulations gives:
\begin{eqnarray}
    &\bDelta^{(k)}(s)=\left(s\bI+\lambda\bH_k\right)^{-1}\left[\bdelta^{(k)}(0)-\frac{a_k}{\sqrt{m}}\bH_k\bDelta(s)\right],\nwl
    &\bDelta(s)=(\bI+\bT(s))^{-1}\suml_k(s\bI+\lambda\bH_k)^{-1}\bdelta^{(k)}(0),
\end{eqnarray}
where $\bDelta^{(k)}(s)$ and $\bDelta(s)$ are respectively the Laplace transforms of $\bdelta^{(k)}(y)$ and $\bdelta(t)$. \\ Hence, $\bdelta(t)$ is given by taking the inverse Laplace transform of $\bDelta(s)$. Note that by construction, $\bDelta(s)$ is a rational function, which shows the finite order of the dynamics. To find the inverse Laplace transform, we only need to find the poles of $\bDelta(s)$. These poles can only be either among the eigenvalues of $\bH_k$ or the values $-p_k$ where the matrix $\bI+\bT(s)$ becomes rank deficient. Under Assumption 1 and 2, we may conclude  that the poles are only $-p_k$, which gives the result in Theorem 1 and 2. More details of this approach can be found in the Section \ref{sec:proofs}, where the kernel hypothesis for this case is also rigorously proved.

\section{Experimental Results}
In this section we present validation for the main results in the paper which helps understanding the theorems and reinforces them. 
We perform our numerical analysis on a commonly-used dataset for validating deep neural models, i.e., CIFAR-10. This dataset is used for the experiments in \cite{Arora19}. As in \cite{Arora19}, we only look at the first two classes and set the label $y_i=+1$ if image $i$ belongs to the first class and $y_i=-1$ if it belongs to the second class. The images $\{x_i\}_{i=1}^{n}$ are normalized such that $||\bm x_i||_2=1$ for all $i = 1, \dots, n$. 
%
The weights in our model are initialized as follows:
\begin{equation}
    \bm w_i \sim \mathcal{N}(0, k^2 \bm{\mathcal{I}}), \, a_r \sim \text{Unif}(\{-1,1\}), \forall r \in [m].
\end{equation}


For optimization, we use (full batch) gradient descent with the learning rate $\eta$. In our experiments we set $k=10^{-2}, \eta=2 \times 10^{-4}$ similar to \cite{Arora19}. 
In all of our experiments we use 100 hidden neurons for the teacher network and 20 hidden neurons for the student network.

\begin{figure}[!htb]
    \centering
    \subfigure[Training loss]
    {
        \includegraphics[width=0.45\columnwidth]{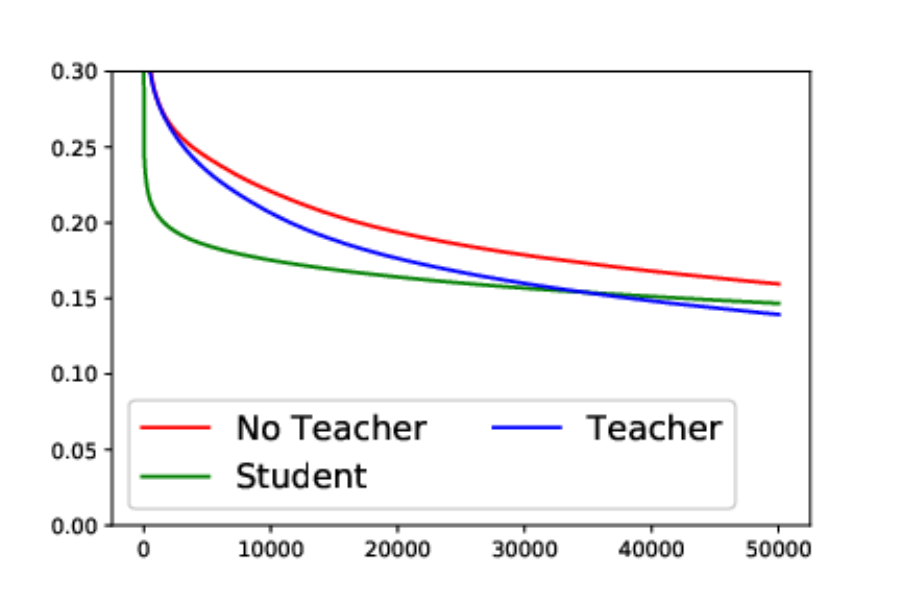}
        \label{fig:train_loss_noteachervstudent}
    }
    \subfigure[Test loss]
    {
        \includegraphics[width=0.45\columnwidth]{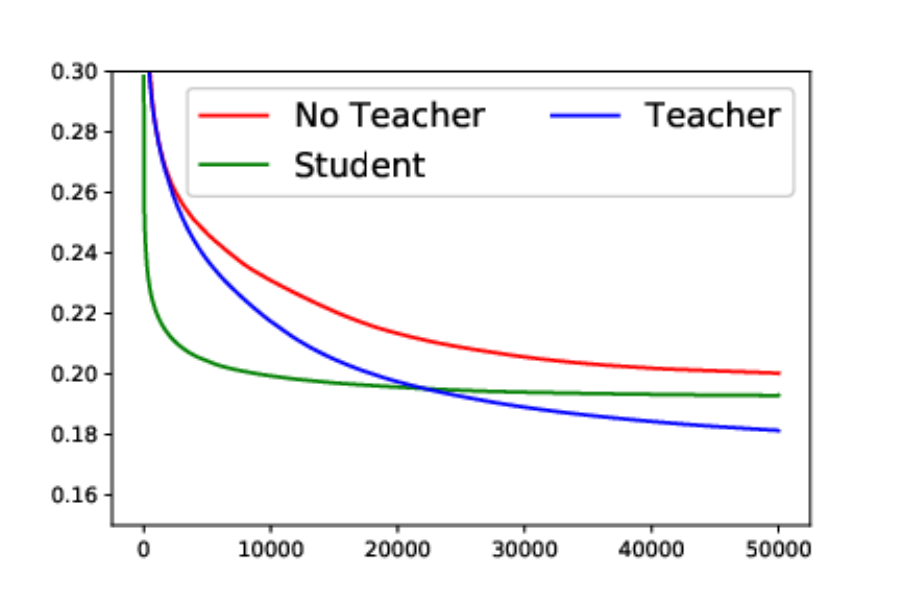}
        \label{fig:testloss_noteachervsstudent}
    }
    \subfigure[Training loss]
    {
        \includegraphics[width=0.45\columnwidth]{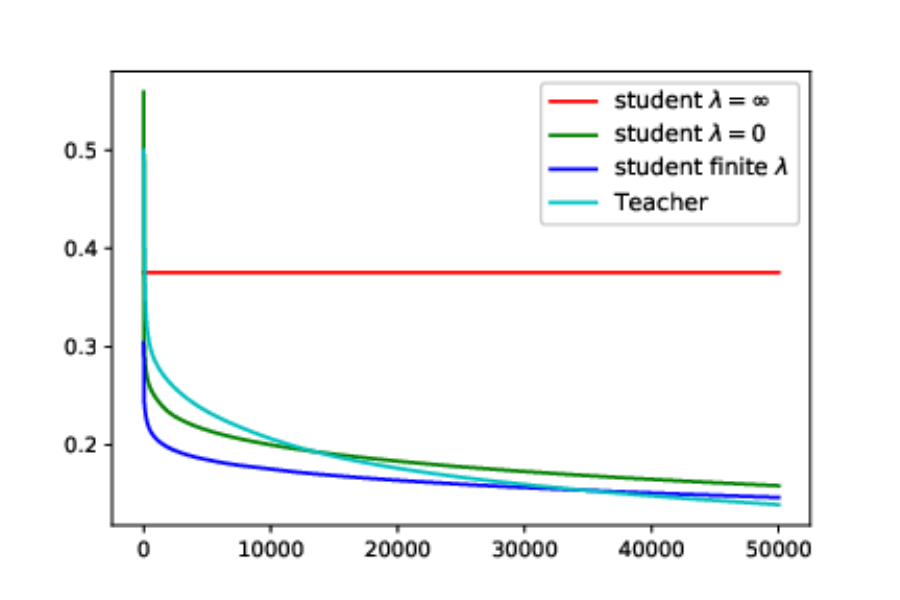}
        \label{fig:testloss_randomvsretrain}
    }
    \subfigure[Test loss]
    {
        \includegraphics[width=0.45\columnwidth]{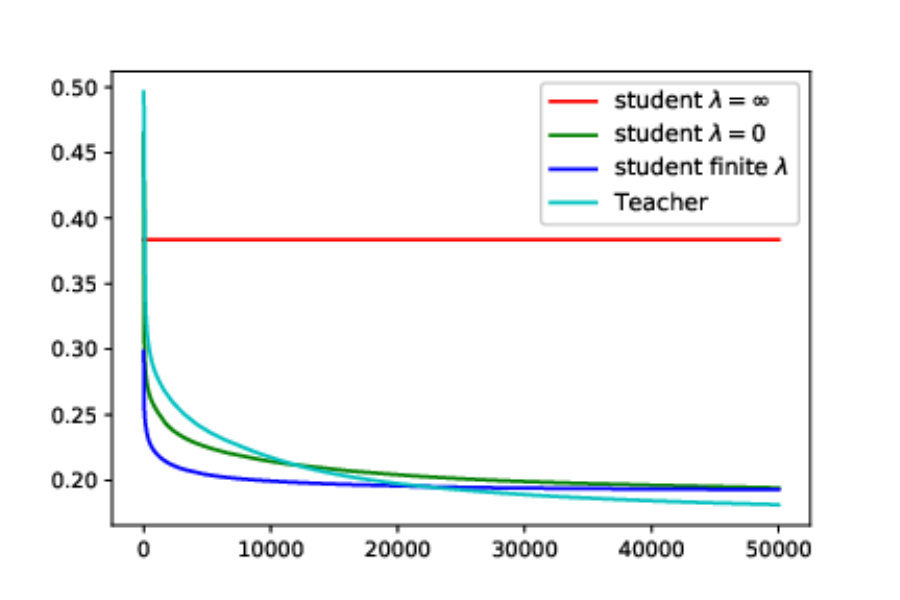}
        \label{fig:test_acc_randomvsretrain}
    }
    \caption{Dynamics of knowledge transfer (a,b) and Effect of different regularization $\lambda$ (c,d).
    }
    \label{fig:KD_dynamics} 
\end{figure}

\subsection{Dynamics of knowledge transfer}
The experiments in this section show the theoretical justification (in Theorem \ref{thm:1} and \ref{thm:3}) of the experimentally well-studied advantage of a teacher. We study knowledge transfer in different settings. We first consider a finite regularization in Eq. \ref{eq:KT_main} by setting $\lambda = 0.01$. 
Figures \ref{fig:train_loss_noteachervstudent} and \ref{fig:testloss_noteachervsstudent} show the dynamics of the  results in different settings, i) no teacher, i.e., the student is independently trained without access to a teacher, ii) student training, where the student is trained by both the teacher and the true labels according to Eq. \ref{eq:KT_main}, and iii) the teacher, trained by only the true labels. For each setting, we illustrate the training loss and the test loss. The horizontal axes in the plots show the number of iterations of the optimizer. In total, we choose 50000 iterations to be sure of the optimization convergence. This corresponds to the parameter $r$ of the dynamical system proposed in section \ref{sec:general_framework}.
Note that true labels are the same for the teacher and the students. Teacher shows the best performance because of its considerably larger capacity. On the other hand, we observe, i) for the student with access to the teacher its performance is better than the student without access to the teacher. This observation is verified by the result in Theorem \ref{thm:1} that states the final performance of the student is a weighted average of the performances of the teacher and of the student with no teacher. This observation is also consistent with the discussion in section \ref{sec:further}, where the final performance of the student is shown to improve with the teacher. ii) The convergence rate of the optimization is significantly faster for the student with teacher compared to the other alternatives. This confirms the prediction of Theorem \ref{thm:3}. This experiment implies the importance of a proper knowledge transfer to the student network via the information from the teacher.

In the following we study the effect of the regularization parameter ($\lambda$) on the dynamics, when a teacher with a similar structure to the student is utilized. The teacher is wider than the student and randomly selected features of the teacher are used as the knowledge $\phi_k$ transferred to the student.  Specifically, we study two special cases of the generic formulation in Eq. \ref{eq:KT_main} where $\lambda \to 0$ and $\lambda \to \infty$. 
Figures  \ref{fig:testloss_randomvsretrain} and \ref{fig:test_acc_randomvsretrain} compare these two extreme cases with the student with $\lambda = 0.01$ and the teacher w.r.t. training loss and test loss. We observe that the student with a finite regularization ($\lambda = 0.01$) outperforms the two other students in terms of both convergence rate (optimization speed) and the quality of the results. In particular, when the student is trained with $\lambda \to \infty$ and it is initialized with the weights of the teacher, then the generic loss in Eq. \ref{eq:KT_main} equals 0. This renders the student network to keep its weights unchanged for $\lambda \to \infty$ and the performance remains equal to that of the privileged knowledge $\suml_k\frac{a_k}{\sqrt{m}}\bphi_k$ without data labels. 

\begin{figure}[!htb]
    \centering
    \subfigure[Training loss]
    {
        \includegraphics[width=0.45\columnwidth]{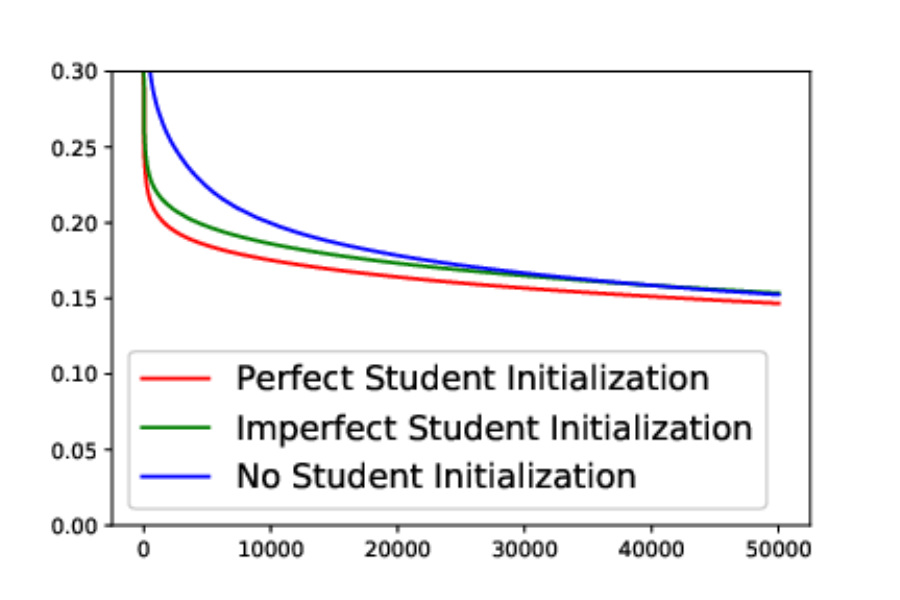}
        \label{fig:train_loss_differentstudents}
    }
    \subfigure[Test loss]
    {
        \includegraphics[width=0.45\columnwidth]{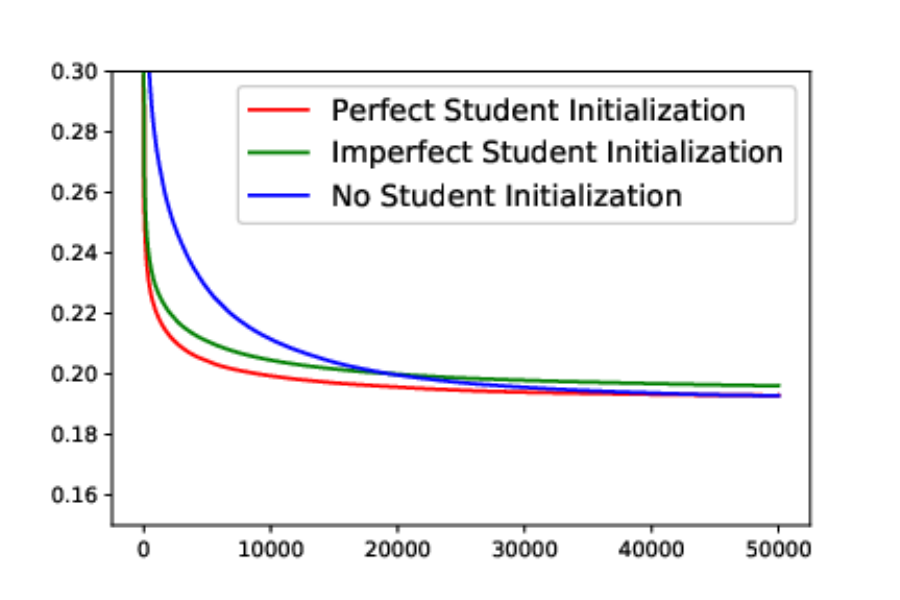}
        \label{fig:testloss_differentstudents}
    }

    \caption{Dynamics of  knowledge transfer with perfect and imperfect teacher. 
}
    \label{fig:Different Teachers}
\end{figure}

\subsection{Dynamics of knowledge transfer with imperfect teacher}
In this section, we study the impact of the quality of the teacher on the student network. We consider the student-teacher scenario in three different settings, i) perfect teacher where the student is initialized with the final weights of the teacher and uses the final teacher outputs in Eq. \ref{eq:KT_main}, ii) imperfect teacher where the student is initialized with the intermediate (early) weights of the teacher network and uses the respective intermediate teacher outputs in Eq. \ref{eq:KT_main}, and iii) no student initialization where the student is initialized randomly but uses the final teacher outputs. In all the settings, we assume $\lambda = 0.01$. 

Figure \ref{fig:Different Teachers} shows the results for these three settings, respectively w.r.t. training loss and test loss.  We observe that initializing and  training the student with the perfect (fully trained) teacher yields the best results in terms of both quality (training and test loss) and convergence rate (optimization speed). 
This observation verifies our theoretical analysis on the importance of initialization of the student with fully trained teacher, as the student should be very close to the teacher.  
\subsection{Kernel embedding}
\label{sec:kernelembedding}
To provide the teacher and the student with more relevant information and to study the role of the data geometry (Theorem \ref{thm:3}), we can use properly designed kernel embeddings. Specifically, instead of using the original features for the networks, we could first learn an optimal kernel which is highly aligned with the labels in training data, implicitly improving the combination of $\{\alpha_j\}$ in Theorem \ref{thm:3} and then we feed the features induced by that kernel embedding into the networks (both student and teacher).

For this purpose, we employ the method proposed in \cite{CortesMohri2012} that develops an algorithm to learn a new kernel from a group of kernels according to a similarity measure between the kernels, namely centered alignment. Then, the problem of learning a kernel with a maximum  alignment between the input data and the labels is formulated as a quadratic programming (QP) problem. The respective algorithm is known as \emph{alignf} \cite{CortesMohri2012}. 

Let us denote by $K^c$ the centered variant of a kernel matrix $K$. 
To obtain the optimal combination of the kernels (i.e., a weighted combination of some base kernels), \cite{CortesMohri2012} suggests the objective function to be centered alignment between the combination of the kernels and $yy^T$, where $y$ is the true labels vector. By restricting the weights to be non-negative, a QP can be formulated as minimizing
 $\hspace{4px} v^TMv - 2v^Ta \hspace{4px}
 w.r.t.\hspace{2px} v \in R_+^P$
where $P$ is the number of the base kernels and $M_{kl} = \langle K_k^c,K_l^c\rangle_F$ for $k,l \in [1,P]$, and finally $a$ is a vector wherein $a_i = \langle K_i^c,yy^T\rangle_F$ for $i \in [1,P]$. If $v^*$ is the solution of the QP, then the vector of kernel weights is given by 
$\mu^* = v^*/\|v^*\|$
 \cite{CortesMohri2012,Gonen2011}.

\begin{figure}[htb!]
    \centering
    \subfigure[Training loss]
    {
        \includegraphics[width=0.45\columnwidth]{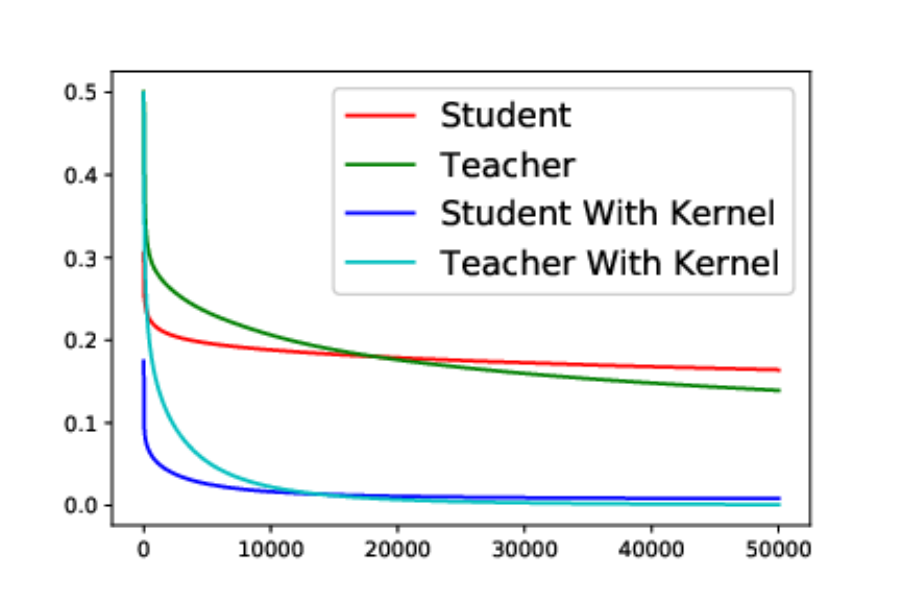}
        \label{fig:train_loss_kernelvsNokernel}
    }
    \subfigure[Test loss]
    {
        \includegraphics[width=0.45\columnwidth]{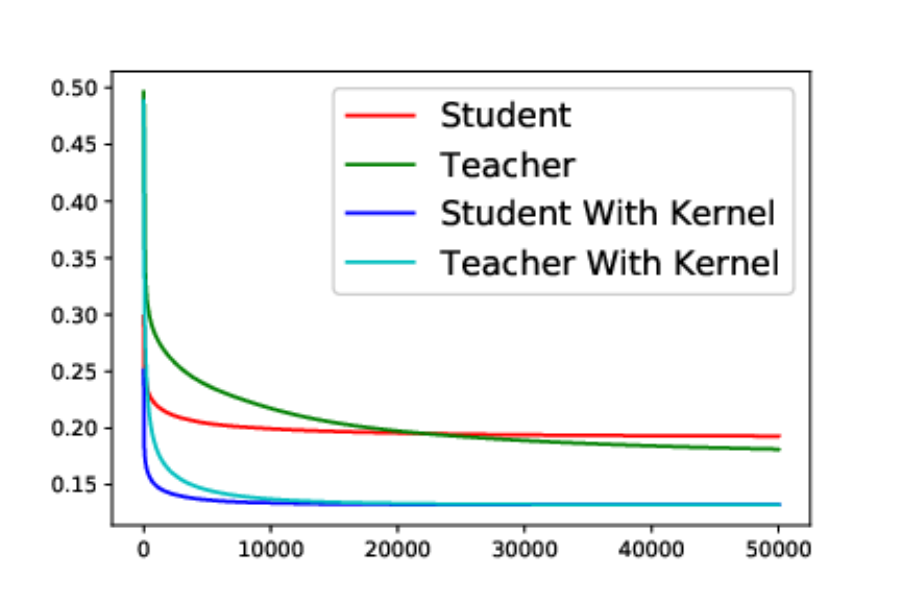}
        \label{fig:test_loss_kernelvsNokernel}
    }

    \caption{Dynamics of knowledge transfer with teacher trained using kernel embedding. 
}
    \label{fig:Kernel}
\end{figure}

Using this algorithm we learn an optimal kernel based on seven different Gaussian kernels. Then, we need to approximate the kernel embeddings. To do so, we use the Nystr\"{o}m method \cite{Nystroem}. Then we feed the approximated embeddings to the neural networks.
The results in Figure \ref{fig:Kernel} show that using the kernel embeddings as inputs to the neural networks, helps both teacher and student networks in terms of training loss (Figure \ref{fig:train_loss_kernelvsNokernel}) and test loss (Figure \ref{fig:test_loss_kernelvsNokernel}).


\subsection{Spectral analysis}
\label{sec:spectralanalysis}
Here, we investigate the overlap parameter of different networks, where we compute a simplified but conceptually consistent variant of the overlap parameter $\alpha_j$ in theorem \ref{thm:3}. For a specific network, we consider the normalized columns of matrix $\phi$ (as defined in Eq. \ref{eq:KT_main}) corresponding to the nonlinear outputs of the hidden neurons, and compute the dot product of each column with the top eigenvectors of $\bH^\infty$, and take the average. We repeat this for all the columns and depict the histogram. For a small value of $\lambda$, the resulting values are approximately equal to $\{\alpha_j\}$ in Theorem \ref{thm:3}.

\begin{figure}[htb!]
    \centering
    \subfigure[Teachers]
    {
        \includegraphics[width=0.45\columnwidth]{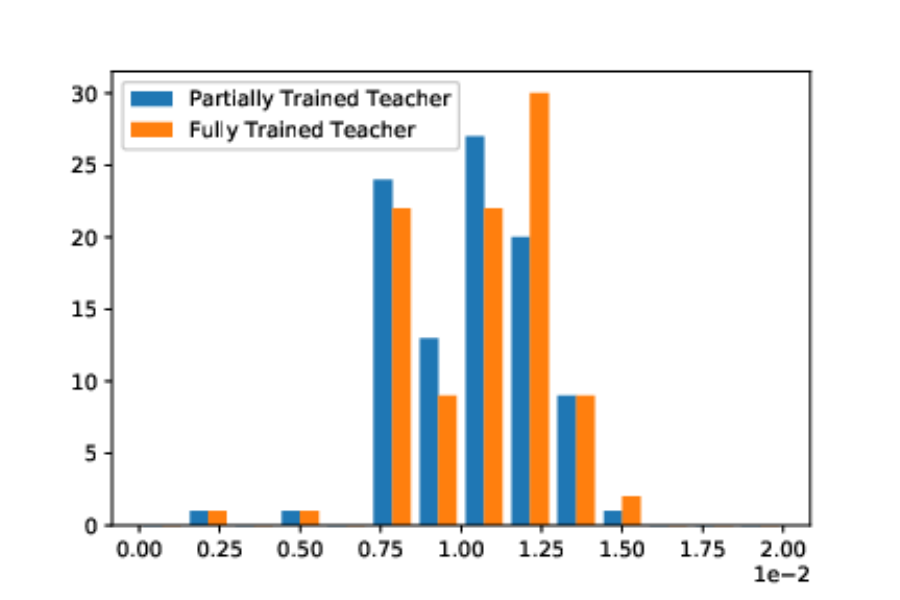}
        \label{fig:spectral_partialvsfully}
    }
    \subfigure[Students]
    {
        \includegraphics[width=0.45\columnwidth]{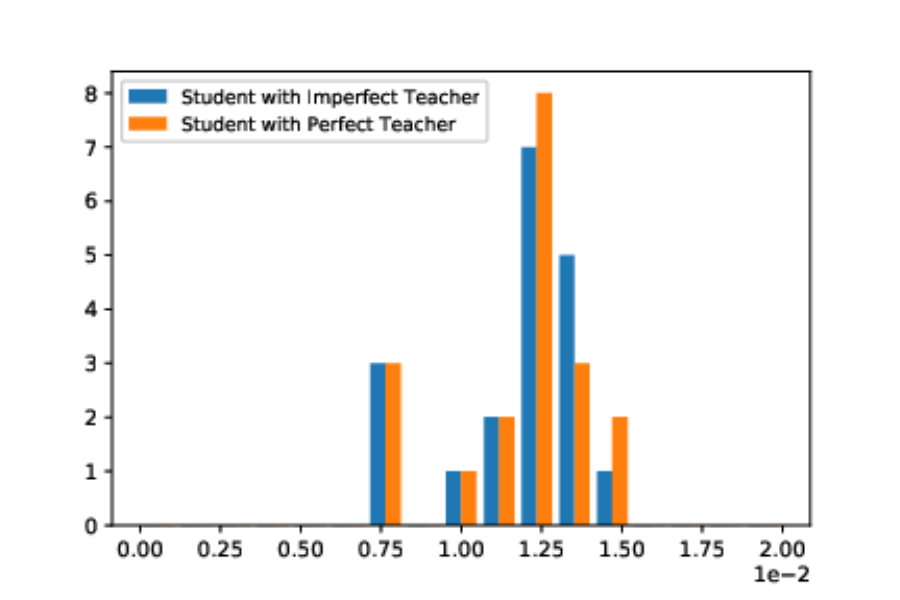}
        \label{fig:spectral_kernel}
    }
    \caption{Spectral analysis of teachers (\ref{fig:spectral_partialvsfully}) and students (\ref{fig:spectral_kernel}) in different settings. 
}
    \label{fig:Different Students}
\end{figure}

Figure~\ref{fig:Different Students} shows such histograms for two settings. In Figure~\ref{fig:spectral_partialvsfully} we compare the overlap parameter for two teachers, one trained partially (imperfect teacher) and the other trained fully (perfect teacher). We observe that the overlap parameter is larger for the teacher trained perfectly, i.e., there is more consistency between its outputs $\phi$ and the matrix $\bH^\infty$. This analysis is consistent with the results in Figure \ref{fig:Different Teachers} which demonstrates the importance of fully trained (perfect) teacher. In Figure~\ref{fig:spectral_kernel}, we show  that this improvement is transferred to the student. 
\section{Proofs}
\label{sec:proofs}




\subsection{Elaborations on Assumptions}
Our analysis will also be built upon a number of assumptions:
\newtheorem{asp}{Assumption}\label{as2}
\begin{asp}

Nonzero eigenvalues of the matrices $\{\bH_k\}$ are all distinct. Note that they are always strictly positive as $\{\bH_k\}$ are by construction positive semi-definite (psd). 
\end{asp}
\begin{asp}
The  values of $p_1,p_2,\ldots,p_d$ are all distinct and different to the eigenvalues of $\{\bH_k\}$.
\end{asp}
\begin{asp}
The function $\sigma$ and its derivative $\sigma^\prime$ are Lipschitz continuous.
\end{asp}
\begin{asp}
We assume $m\to\infty$, such that $\|\bphi_k-\mbf_k(0)\|=O(1/\sqrt{m})$ and $\|\by-\suml_k\frac{\ba_k\phi_k}{\sqrt m}\|=O(1)$.
\end{asp}
\begin{asp}
$\|\bx_i\|$s and $a_k$s are bounded.
\end{asp}
\begin{asp}
$\suml_k\frac{\bara_k^2}{\barm}\|\barbphi_k\|^2$ is bounded.
\end{asp}
Assumption 1-5 are required for Theorem 1. Assumption 1-6 are required for Theorem 3.

In practice, these assumptions are mild, as we elaborate in the following:\\
    1. Assumption 1 and 2 are equivalent to assuming that all eigenvalues of $\barbH$, defined in the concluding lines of section \ref{sec:dyn}, are distinct. Note that $\barbH$ depends on the data and hence inherits its random nature. Accordingly, the event that such a random matrix has an eigenvalue with multiplicity is “zero-measure”, i.e almost impossible in reality.\\
    2. Assumption 3 holds for virtually every popular activation functions, including ReLU and Sigmoid.\\
    3. Assumption 4 is equivalent to the scenario in the experiments of section \ref{sec:kernelembedding}, referred to as \emph{almost perfect initialization of student by teacher}. Note that it can be rather viewed as a design specification and our experiments verify its merits. The other experiments of section \ref{sec:kernelembedding} identify another consistent (but still weaker) scenario, beyond this assumption. \\
    4. Assumptions 5 can be easily be imposed by normalizing the data. Similar assumptions also exist in \cite{du2018gradient,Arora19}, but ours are slightly weaker. for example, they merely assume binary values for $a_k$.\\
    5. Assumption 6 is also very mild. For the student-teacher scenario for example, it requires the features of the teacher to be bounded in the $\ell_2$ sense, which is met by standard network architectures. 

\subsection{Proof of Theorem 1 and 2}
\label{thm13proof}

 We continue the discussion in (12 of paper) and remind that $\sigma_{\max}$ is the largest singular value of matrices and $\|\|$ means the 2-norm of vectors. Note that the values $s=-p_i$ correspond to the points where $\det(\bI+\bT(s))=0$. We also observe that these values correspond to the negative of eigenvalues of the matrix $\barbH(t=0)$. We conclude that under assumption 2, the eigenvalues of  $\barbH=\barbH(t=0)$ are distinct and strictly positive, hence this matrix is diagonalizable. Now, we write $\barbH(t)=\barbH+\Delta\barbH(t)$ and state the following lemma:
\begin{lemma}
Suppose that $\barbH$ is a diagonalizable matrix with strictly positive eigenvalues and denote its smallest eigenvalue by $p$. Take $\Delta\barbH(t)$ as a matrix valued function of the continuous valiable $t$ such that for a given fixed value of $t$
\[
q=q(t)=\frac{\supl_{\tau\in [0\ t]}\sigma_{\max}(\Delta\barbH(\tau))}{p}<1
\]
Let $\bmeta(t)$ denote the solution to $\nicefrac{\td\bmeta}{\td t}(t)=-\barbH(t)\bmeta(t),$ with $\barbH(t)=\barbH+\Delta\barbH(t)$. Then, 
\[
\intl_0^t\left\|\bmeta(\tau)-e^{-\barbH \tau}\bmeta(0)\right\|\td\tau\leq\frac{q\|\bmeta(0)\|}{p(1-q)}
\]
\begin{proof}
Consider the iteration $\bmeta^{r+1}=\calT\bmeta^{r}$ that generates a sequence of function functions $\bmeta^{r}(t)$ for $r=0,1,\ldots$ where $\bmeta^{r}(0)=e^{-\barbH t}\bmeta(0)$ and $\bmeta^\prime=\calT\bmeta$ is the solution to 
\begin{equation}
    \frac{\td}{\td t}\bmeta^\prime(t)=-\barbH\bmeta^\prime(t)-\Delta\barbH(t)\bmeta(t)
\end{equation}
with $\bmeta^\prime(0)=\bmeta(0)$, which can also be written as
\begin{equation}
    \bmeta^\prime(t)=e^{-\barbH t}\bmeta(0)-\intl_0^{t}e^{-\barbH (t-\tau)}\Delta\barbH(\tau)\bmeta(\tau)\td\tau
\end{equation}
We observe that $\calT$ on the interval $[0\ t]$ is a contraction map under $L_1$ norm as we have
\begin{equation}
    \Delta\bmeta^\prime(t)=-\intl_0^{t}e^{-\barbH (t-\tau)}\Delta\barbH(\tau)\Delta\bmeta(\tau)\td\tau
\end{equation}
and hence by the triangle inequality , we get
\begin{eqnarray}
    &\|\Delta\bmeta^\prime(t)\|_2\leq\supl_{\tau\in[0\ t]}\sigma_{\max}(\Delta\barbH(\tau))\times\nwl
    &\intl_0^{t}e^{-p(t-\tau)}\|\Delta\bmeta(\tau)\|_2\td\tau 
\end{eqnarray}
we conclude that
\[
\intl_0^t\|\Delta\bmeta^\prime(\tau)\|_2\td t\leq 
\]
\[
\supl_{\tau\in[0\ t]}\sigma_{\max}(\Delta\barbH(\tau))\times \intl_0^t\frac{1-e^{-p(t-\tau)}}{p}\|\Delta\bmeta(\tau)\|_2
\]
\[
\leq q\intl_0^t\|\Delta\bmeta(\tau)\|_2\td t
\]
which shows that $\calT$ is a contraction. Then, from Banach fixed-point theorem we conclude that $\bmeta^{r}$ converges uniformly on the interval $[0\ t]$ to the fixed-point $\bmeta$ of $\calT$, which coincides with the solution of (14 in paper). Moreover,
\[
\intl_0^t\|\bmeta-\bmeta^0(\tau)\|_2\td\tau\leq \frac{\intl_0^t\|\bmeta^1(\tau)-\bmeta^0(\tau)\|_2\td\tau}{1-q}
\]
Now, we observe that
\[
\bmeta^1(t)-\bmeta^0(t)=-\intl_0^te^{-\barbH(t-\tau)}\Delta\barbH(\tau)e^{-\barbH(\tau)}\bmeta(0)\td\tau
\]
Hence,
\[
\|\bmeta^1(t)-\bmeta^0(t)\|\leq \intl_0^te^{-p(t-\tau)}\sigma_{\max}(\Delta\barbH(\tau))e^{-p\tau}\|\bmeta(0)\|\td\tau
\]
\[
\leq te^{-pt}\supl_{\tau\in [0\ t]}\sigma_{\max}(\Delta\barbH(\tau))\|\bmeta(0)\|
\]
and
\[
\intl_0^t\|\bmeta^1(\tau)-\bmeta^0(\tau)\|\td\tau\leq
\]
\[
 \supl_{\tau\in [0\ t]}\sigma_{\max}(\Delta\barbH(\tau))\|\bmeta(0)\|\intl_0^t\tau e^{-p\tau}\td\tau\leq\frac q p\|\bmeta(0)\|
\]
which completes the proof.
\end{proof}
\end{lemma}
Now, we state two results that connect $\sigma_{\max}(\bDelta\bH)$ to the change of $\bw_k(t)$:

\begin{lemma}
Under Assumption 3, the following relation holds:
\begin{equation}
    \sigma_{\max}(\barbH(t))\leq\sqrt 2 \sqrt{\lambda^2+\left(\suml_k\frac{a_k^2}{m}\right)^2}\maxl_k\sigma_{\max}(\Delta\bH_k(t))
\end{equation}
where $\Delta\bH_k(t)=\bH_k(t)-\bH_k(0)$.
\begin{proof}
Take an arbitrary block vector $\bmeta=[\bdelta_k]_k$ with $\|\bmeta\|=1$ and note that
\[
\|\Delta\barbH(t)\bmeta\|^2=\suml_k\|\Delta\bH_k(t)(\frac{a_k}{\sqrt{m}}\bdelta+\lambda\bdelta_k)\|^2
\]
\[
\leq 2\maxl_k\sigma^2_{\max}(\Delta\bH_k(t))\suml_k\left(\frac{a_k^2}m\|\bdelta\|^2+\lambda\|\bdelta_k\|^2\right)
\]
\[
=2\maxl_k\sigma^2_{\max}(\Delta\bH_k(t))\left(\|\bdelta\|^2\suml_k\frac{a_k^2}m+\lambda^2\right),
\]
where $\bdelta=\suml_k\frac{a_k}{\sqrt{m}}\bdelta_k$. We obtain the desired result by observing that
\[
\|\bdelta\|^2=\left\|\suml_k\frac{a_k}{\sqrt{m}}\bdelta_k\right\|^2\leq\left(\suml_k\frac{a_k^2}{m}\right)\suml_k\|\bdelta_k\|^2
=\suml_k\frac{a_k^2}{m}.
\]
\end{proof}

\end{lemma}
Next, we show
\begin{lemma}
We have
\[
\sigma_{\max}(\Delta\bH_k(t))\leq L^2\sigma^2_x\maxl_i\|\bx_i\|^2\times 
\]
\[
\|\bw_k(t)-\bw_k(0)\|(\|\bw_k(t)-\bw_k(0)\|+2\|\bw_k(0)\|)
\]
where $\sigma_x$ is the maximal eigenvalue of the the data matrix $\bX=[\bx_1\ \bx_2\ldots \bx_n]$ and $L$ is the largest of the Lipschitz constants of $\sigma,\sigma^\prime$.
\begin{proof}
Note that since $\Delta\bH_k$ is symmetric, we have (e.g. by eigen-decomposition)
\[
\sigma_{\max}(\Delta\bH_k(t))=\max_{\bdelta\mid\|\bdelta\|=1}|\bdelta^T\Delta\bH_k(t)\bdelta|
\]
Taking an arbitrary normalized $\bdelta$, we observe that
\[
\bdelta^T\Delta\bH_k(t)\bdelta=\left|\|\bL(\bw_k(t))\bdelta\|^2-\|\bL(\bw_k(0))\bdelta\|^2\right|
\]
On other hand,
\[
\bL(\bw_k(t))\bdelta=\bL(\bw_k(0))\bdelta+
\suml_i\bx_i\sigma_i\delta_i,
\]
where $\sigma_i=\sigma^\prime(\bw_k(t)^T\bx_i)-\sigma^\prime(\bw_k(0)^T\bx_i)$. Hence, 
\[
\bdelta^T\Delta\bH_k(t)\bdelta\leq \left\|\suml_i\bx_i\sigma_i\delta_i\right\|^2 + 2\left\|\suml_i\bx_i\delta_i\sigma_i\right\|\|\bL(\bw_k(0))\bdelta\|
\]
We also observe that
\[
\left\|\suml_i\bx_i\sigma_i\delta_i\right\|\leq\sigma_x\sqrt{\suml_i\delta_i^2\sigma_i^2}
\]
and from Lipschitz continuity, 
\[
\sigma_i^2\leq L^2\langle\bx_i,\bw_k(t)-\bw_k(0)\rangle^2\leq L^2\|\bx_i\|^2\|\bw_k(t)-\bw_k(0)\|^2
\]
We conclude that
\[
\left\|\suml_i\bx_i\sigma_i\delta_i\right\|\leq L\sigma_x\|\bw_k(t)-\bw_k(0)\|\times\maxl_i\|\bx_i\|
\]
Similarly, we obtain
\[
\left\|\bL(\bw_k(0))\bdelta\right\|\leq L\sigma_x\times\maxl_i\|\bx_i\|
\]
which completes the proof.
\end{proof}
\end{lemma}

We finally connect the  magnitude of the change $\bw_k(t)-\bw_k(0)$ to $\bdelta_k$:
\begin{lemma}
 With the same definitions as in Lemma 3, we have
 \begin{equation}
     \|\bw_k(t)-\bw_k(0)\|\leq\frac{|a_k|}{\sqrt m} L\sigma_x\maxl_i\|\bx_i\|\intl_0^t\|\bdelta_k(\tau)\|\td\tau
 \end{equation}
 \begin{proof}
 Note that
 $
 \bw_k(t)-\bw_k(0)=-\frac{a_k}{\sqrt m}\intl_0^t\bL(\bw_k(\tau))\bdelta_k(\tau)\td\tau
 $
 and hence
 $
 \|\bw_k(t)-\bw_k(0)\|\leq\frac{|a_k|}{\sqrt m}\intl_0^t\|\bL(\bw_k(\tau))\bdelta_k(\tau)\|\td\tau
 $
 From a similar argument as in Lemma 3, we have
 \[
 \|\bL(\bw_k(\tau))\bdelta_k(\tau)\|\leq L\sigma_x\maxl_i\|\bx_i\|\|\bdelta_k(\tau)\|
 \]
 which completes the proof.
 \end{proof}
\end{lemma}
We may now proceed to the proof of Theorem 1 and 2. Define
\[
T=\{t\mid\forall\tau\in [0\ t];\ q(\tau)<\frac 12\}
\]
Note that $T$ is nonempty as $0\in T$ and open since $q$ is continuous. We show that for sufficiently large $m$, $T=[0,\infty)$. Otherwise $T$ is an open interval $[0\ t_0)$ where $q(t_0)=\frac 1 2$. For any $t\in T$, we have from Lemma 1
\[
A=\intl_0^t\|\bmeta(\tau)\|\td\tau\leq\left(\frac{1-e^{-pt}}{p}+\frac{q}{p(1-q)}\right)\|\bmeta(0)\|
\leq \frac{\|\bmeta(0)\|}{p(1-q)}
\]
Denote $b=\maxl_k|a_k|$ and $B=\maxl_i\|\bx_i\|$. We further define 
$
C=\frac{\sqrt 2}pL^3\sigma_B^3x^3b\sqrt{\lambda^2+a^2}.
$
Then Lemma 2,3 and 4 give us 
$
q\leq\frac{CA}{\sqrt m}.
$
We conclude that
\[
q(t)\leq \frac{C\|\bmeta(0)\|}{p(1-q(t))\sqrt m}
\]
Note that by Assumption 4, we have
\[
\bdelta_k(0)=\frac{a_k}{(\lambda+a)\sqrt m}(\by-\suml_k\frac{a_k\bphi_k}{\sqrt m})
+\bphi_k-\mbf_k(0)=O(1/\sqrt m)
\]
and hence $\bmeta(0)=O(1)$.  This shows that there exists a constant $C_0$ such that
$
q(t)(1-q(t))\leq\frac{C_0}{\sqrt m}
$
for $t\in T$. But for large values of $m$ this is in contradiction to $q(t_0)=1/2$. Hence, for such values $t_0$ does not exist and $q(t)<1/2$ for all $t$. We conclude that for sufficiently large values of $m$ we have $q(t)<3C_0/\sqrt{m}$ for all $t$. Then, according to Lemma 1 and the monotone convergence theorem, we have
\[
\intl_0^\infty\left\|\bmeta(\tau)-e^{-\barbH \tau}\bmeta(0)\right\|\td\tau=O(\frac 1{\sqrt m})
\]
This shows that
\[
\liml_{t\to\infty}\|\bmeta(t)-e^{-\barbH t}\bmeta(0)\|=0
\]
Note that as $\barbH$ is diagonalizable and has strictly positive eigenvalues, we get that 
\[
\liml_{t\to\infty}e^{-\barbH t}\bmeta(0)=0,
\]
which further leads to 
\[
\liml_{t\to\infty}\bmeta(t)=0,
\]
This proves Theorem 1. For Theorem 2, we see that
\[
\bmeta(t)=e^{-\barbH t}\bmeta(0)+O(1/\sqrt m)
\]
It suffices to show that the expression in Theorem 2 coincides with $e^{-\barbH t}\bmeta(0)$. This is simple to see through the following lemma:

\begin{lemma}
According to Assumption 2, the right and left eigenvectors of $\barbH$ corresponding to $p_j$ are respectively given by vectors $\bmeta^{\mathrm{r}}_j=[\bv^j_k]_k$ and  $\bmeta^{\mathrm{l}}_j=[\bu^j_k]_k$, where
$
\bv_k^j=\frac{a_k}{\sqrt m}(p_j\bI-\lambda\bH_k)^{-1}\bH_k\bv^j
$
and
$
\bu_k^j=\frac{a_k}{\sqrt m}(p_j\bI-\lambda\bH_k)^{-1}\bu^j.
$
Moreover, $\bv^j=\suml_k\frac{a_k}{\sqrt m}\bv^j_k$.
\begin{proof}
According to the definition of $\barbH$, we have that
\[
\bH_k(\frac{a_k}{\sqrt m}\bv+\lambda\bv^j_k)=p\bv^j_k
\]
where $\bv=\suml_k\frac{a_k}{\sqrt{m}}\bv^j_k$ which gives
$
\bv_k^j=\frac{a_k}{\sqrt m}(p_j\bI-\lambda\bH_k)^{-1}\bH_k\bv
$
Replacing this expression in the definition of $\bv$ shows that $\bv=\bv^j$. The case for $\bu^j_k$ is similarly proved.
\end{proof}
\end{lemma}

Theorem 2 simply follows by replacing the result of Lemma 5 to the eigen-decomposition of $e^{-\barbH t}$:
\[
e^{-\barbH t}=\suml_je^{-p_j  t}\mid\bmeta^{\mathrm{r}}_j\rangle\langle\bmeta^{\mathrm{l}}_j\mid
\]
\subsection{Proof of Theorem 3}
Note that for $p=p_j$ there exists an eigen vector $\bv$ of $\bT(-p)$ such that $\bv^T\bT(-p)\bv=-1$. This leads to
$
-1=\suml_k\frac{a_k^2}m\bv^T\left(-p\bI+\lambda\bH_k\right)\bH_k\bv.
$
If the result does not hold, we have that $-p\bI+\lambda\bH_k\succeq \bzero$. Hence, the right hands side is non-negative, as $\bH_k$ is also psd and the two matrices commute, leading to a contradiction. This proves the result.
\section{Conclusions}
We give a theoretical analysis of knowledge transfer for non--linear neural networks in the model and regime of \cite{Arora19,DuH19,CaoG19a} which yields insights on both privileged information and knowledge distillation paradigms.
We provide results for both what is learnt by the student and on the speed of convergence. We further provide a discussion about the effect of knowledge transfer on generalization.
Our numerical studies further confirm our theoretical findings on the role of data geometry and knowledge transfer in the final performance of student.

\bibliographystyle{ACM-Reference-Format}
\balance
\bibliography{refs}


\begin{thebibliography}{22}


\ifx \showCODEN    \undefined \def \showCODEN     #1{\unskip}     \fi
\ifx \showDOI      \undefined \def \showDOI       #1{#1}\fi
\ifx \showISBNx    \undefined \def \showISBNx     #1{\unskip}     \fi
\ifx \showISBNxiii \undefined \def \showISBNxiii  #1{\unskip}     \fi
\ifx \showISSN     \undefined \def \showISSN      #1{\unskip}     \fi
\ifx \showLCCN     \undefined \def \showLCCN      #1{\unskip}     \fi
\ifx \shownote     \undefined \def \shownote      #1{#1}          \fi
\ifx \showarticletitle \undefined \def \showarticletitle #1{#1}   \fi
\ifx \showURL      \undefined \def \showURL       {\relax}        \fi
\providecommand\bibfield[2]{#2}
\providecommand\bibinfo[2]{#2}
\providecommand\natexlab[1]{#1}
\providecommand\showeprint[2][]{arXiv:#2}

\bibitem[Arora et~al\mbox{.}(2019)]%
        {Arora19}
\bibfield{author}{\bibinfo{person}{Sanjeev Arora}, \bibinfo{person}{Simon~S.
  Du}, \bibinfo{person}{Wei Hu}, \bibinfo{person}{Zhiyuan Li}, {and}
  \bibinfo{person}{Ruosong Wang}.} \bibinfo{year}{2019}\natexlab{}.
\newblock \showarticletitle{Fine-Grained Analysis of Optimization and
  Generalization for Overparameterized Two-Layer Neural Networks}. In
  \bibinfo{booktitle}{\emph{Proceedings of the 36th International Conference on
  Machine Learning, {ICML} 2019, 9-15 June 2019, Long Beach, California,
  {USA}}}. \bibinfo{pages}{322--332}.
\newblock


\bibitem[Cao and Gu(2019)]%
        {CaoG19a}
\bibfield{author}{\bibinfo{person}{Yuan Cao} {and} \bibinfo{person}{Quanquan
  Gu}.} \bibinfo{year}{2019}\natexlab{}.
\newblock \showarticletitle{Generalization Bounds of Stochastic Gradient
  Descent for Wide and Deep Neural Networks}. In
  \bibinfo{booktitle}{\emph{Advances in Neural Information Processing Systems
  32: Annual Conference on Neural Information Processing Systems 2019, NeurIPS
  2019, 8-14 December 2019, Vancouver, BC, Canada}}.
  \bibinfo{pages}{10835--10845}.
\newblock


\bibitem[Chen et~al\mbox{.}(2017)]%
        {chen2017learning}
\bibfield{author}{\bibinfo{person}{Guobin Chen}, \bibinfo{person}{Wongun Choi},
  \bibinfo{person}{Xiang Yu}, \bibinfo{person}{Tony Han}, {and}
  \bibinfo{person}{Manmohan Chandraker}.} \bibinfo{year}{2017}\natexlab{}.
\newblock \showarticletitle{Learning efficient object detection models with
  knowledge distillation}. In \bibinfo{booktitle}{\emph{Advances in Neural
  Information Processing Systems}}. \bibinfo{pages}{742--751}.
\newblock


\bibitem[Cortes et~al\mbox{.}(2012)]%
        {CortesMohri2012}
\bibfield{author}{\bibinfo{person}{Corinna Cortes}, \bibinfo{person}{Mehryar
  Mohri}, {and} \bibinfo{person}{Afshin Rostamizadeh}.}
  \bibinfo{year}{2012}\natexlab{}.
\newblock \showarticletitle{Algorithms for Learning Kernels Based on Centered
  Alignment}.
\newblock \bibinfo{journal}{\emph{J. Mach. Learn. Res.}}  \bibinfo{volume}{13}
  (\bibinfo{date}{March} \bibinfo{year}{2012}), \bibinfo{pages}{795--828}.
\newblock
\showISSN{1532-4435}


\bibitem[Du and Hu(2019)]%
        {DuH19}
\bibfield{author}{\bibinfo{person}{Simon~S. Du} {and} \bibinfo{person}{Wei
  Hu}.} \bibinfo{year}{2019}\natexlab{}.
\newblock \showarticletitle{Width Provably Matters in Optimization for Deep
  Linear Neural Networks}. In \bibinfo{booktitle}{\emph{Proceedings of the 36th
  International Conference on Machine Learning, {ICML} 2019, 9-15 June 2019,
  Long Beach, California, {USA}}}. \bibinfo{pages}{1655--1664}.
\newblock


\bibitem[Du et~al\mbox{.}(2018)]%
        {du2018gradient}
\bibfield{author}{\bibinfo{person}{Simon~S Du}, \bibinfo{person}{Xiyu Zhai},
  \bibinfo{person}{Barnabas Poczos}, {and} \bibinfo{person}{Aarti Singh}.}
  \bibinfo{year}{2018}\natexlab{}.
\newblock \showarticletitle{Gradient descent provably optimizes
  over-parameterized neural networks}.
\newblock \bibinfo{journal}{\emph{arXiv preprint arXiv:1810.02054}}
  (\bibinfo{year}{2018}).
\newblock


\bibitem[G\"{o}nen and Alpayd(2011)]%
        {Gonen2011}
\bibfield{author}{\bibinfo{person}{Mehmet G\"{o}nen} {and}
  \bibinfo{person}{Ethem Alpayd}.} \bibinfo{year}{2011}\natexlab{}.
\newblock \showarticletitle{Multiple Kernel Learning Algorithms}.
\newblock \bibinfo{journal}{\emph{J. Mach. Learn. Res.}}  \bibinfo{volume}{12}
  (\bibinfo{date}{July} \bibinfo{year}{2011}), \bibinfo{pages}{2211--2268}.
\newblock
\showISSN{1532-4435}


\bibitem[Hinton et~al\mbox{.}(2015)]%
        {hinton15}
\bibfield{author}{\bibinfo{person}{Geoffrey~E. Hinton}, \bibinfo{person}{Oriol
  Vinyals}, {and} \bibinfo{person}{Jeffrey Dean}.}
  \bibinfo{year}{2015}\natexlab{}.
\newblock \showarticletitle{Distilling the Knowledge in a Neural Network}.
\newblock \bibinfo{journal}{\emph{CoRR}}  \bibinfo{volume}{abs/1503.02531}
  (\bibinfo{year}{2015}).
\newblock


\bibitem[Jacot et~al\mbox{.}(2018)]%
        {JFH18}
\bibfield{author}{\bibinfo{person}{Arthur Jacot}, \bibinfo{person}{Franck
  Gabriel}, {and} \bibinfo{person}{Clement Hongler}.}
  \bibinfo{year}{2018}\natexlab{}.
\newblock \showarticletitle{Neural Tangent Kernel: Convergence and
  Generalization in Neural Networks}.
\newblock In \bibinfo{booktitle}{\emph{Advances in Neural Information
  Processing Systems 31}}, \bibfield{editor}{\bibinfo{person}{S.~Bengio},
  \bibinfo{person}{H.~Wallach}, \bibinfo{person}{H.~Larochelle},
  \bibinfo{person}{K.~Grauman}, \bibinfo{person}{N.~Cesa-Bianchi}, {and}
  \bibinfo{person}{R.~Garnett}} (Eds.). \bibinfo{publisher}{Curran Associates,
  Inc.}, \bibinfo{pages}{8571--8580}.
\newblock


\bibitem[Kim and Rush(2016)]%
        {KD-sequence}
\bibfield{author}{\bibinfo{person}{Yoon Kim} {and}
  \bibinfo{person}{Alexander~M. Rush}.} \bibinfo{year}{2016}\natexlab{}.
\newblock \showarticletitle{Sequence-Level Knowledge Distillation}. In
  \bibinfo{booktitle}{\emph{Proceedings of the 2016 Conference on Empirical
  Methods in Natural Language Processing}}. \bibinfo{publisher}{Association for
  Computational Linguistics}, \bibinfo{address}{Austin, Texas},
  \bibinfo{pages}{1317--1327}.
\newblock
\urldef\tempurl%
\url{https://doi.org/10.18653/v1/D16-1139}
\showDOI{\tempurl}


\bibitem[Lopez{-}Paz et~al\mbox{.}(2016)]%
        {LopezPaz15}
\bibfield{author}{\bibinfo{person}{David Lopez{-}Paz},
  \bibinfo{person}{L{\'{e}}on Bottou}, \bibinfo{person}{Bernhard
  Sch{\"{o}}lkopf}, {and} \bibinfo{person}{Vladimir Vapnik}.}
  \bibinfo{year}{2016}\natexlab{}.
\newblock \showarticletitle{Unifying distillation and privileged information}.
  In \bibinfo{booktitle}{\emph{4th International Conference on Learning
  Representations, {ICLR} 2016, San Juan, Puerto Rico, May 2-4, 2016,
  Conference Track Proceedings}}.
\newblock


\bibitem[Mei et~al\mbox{.}(2019)]%
        {mei2019mean}
\bibfield{author}{\bibinfo{person}{Song Mei}, \bibinfo{person}{Theodor
  Misiakiewicz}, {and} \bibinfo{person}{Andrea Montanari}.}
  \bibinfo{year}{2019}\natexlab{}.
\newblock \showarticletitle{Mean-field theory of two-layers neural networks:
  dimension-free bounds and kernel limit}.
\newblock \bibinfo{journal}{\emph{arXiv preprint arXiv:1902.06015}}
  (\bibinfo{year}{2019}).
\newblock


\bibitem[Pan et~al\mbox{.}(2019)]%
        {auto-KD}
\bibfield{author}{\bibinfo{person}{Yiteng Pan}, \bibinfo{person}{Fazhi He},
  {and} \bibinfo{person}{Haiping Yu}.} \bibinfo{year}{2019}\natexlab{}.
\newblock \showarticletitle{A novel Enhanced Collaborative Autoencoder with
  knowledge distillation for top-N recommender systems}.
\newblock \bibinfo{journal}{\emph{Neurocomputing}}  \bibinfo{volume}{332}
  (\bibinfo{year}{2019}), \bibinfo{pages}{137--148}.
\newblock
\showISSN{0925-2312}
\urldef\tempurl%
\url{https://doi.org/10.1016/j.neucom.2018.12.025}
\showDOI{\tempurl}


\bibitem[Pechyony and Vapnik(2010)]%
        {privilegeNips}
\bibfield{author}{\bibinfo{person}{Dmitry Pechyony} {and}
  \bibinfo{person}{Vladimir Vapnik}.} \bibinfo{year}{2010}\natexlab{}.
\newblock \showarticletitle{On the Theory of Learning with Privileged
  Information}. In \bibinfo{booktitle}{\emph{Proceedings of the 23rd
  International Conference on Neural Information Processing Systems - Volume
  2}} (Vancouver, British Columbia, Canada) \emph{(\bibinfo{series}{NIPS'10})}.
  \bibinfo{publisher}{Curran Associates Inc.}, \bibinfo{address}{Red Hook, NY,
  USA}, \bibinfo{pages}{1894–1902}.
\newblock


\bibitem[Phuong and Lampert(2019)]%
        {phuong19}
\bibfield{author}{\bibinfo{person}{Mary Phuong} {and}
  \bibinfo{person}{Christoph Lampert}.} \bibinfo{year}{2019}\natexlab{}.
\newblock \showarticletitle{Towards Understanding Knowledge Distillation}. In
  \bibinfo{booktitle}{\emph{Proceedings of the 36th International Conference on
  Machine Learning}} \emph{(\bibinfo{series}{Proceedings of Machine Learning
  Research}, Vol.~\bibinfo{volume}{97})},
  \bibfield{editor}{\bibinfo{person}{Kamalika Chaudhuri} {and}
  \bibinfo{person}{Ruslan Salakhutdinov}} (Eds.). \bibinfo{publisher}{PMLR},
  \bibinfo{address}{Long Beach, California, USA}, \bibinfo{pages}{5142--5151}.
\newblock


\bibitem[Vapnik and Izmailov(2015)]%
        {JMLRPrivileged}
\bibfield{author}{\bibinfo{person}{Vladimir Vapnik} {and} \bibinfo{person}{Rauf
  Izmailov}.} \bibinfo{year}{2015}\natexlab{}.
\newblock \showarticletitle{Learning Using Privileged Information: Similarity
  Control and Knowledge Transfer}.
\newblock \bibinfo{journal}{\emph{Journal of Machine Learning Research}}
  \bibinfo{volume}{16}, \bibinfo{number}{61} (\bibinfo{year}{2015}),
  \bibinfo{pages}{2023--2049}.
\newblock
\urldef\tempurl%
\url{http://jmlr.org/papers/v16/vapnik15b.html}
\showURL{%
\tempurl}


\bibitem[Vapnik and Izmailov(2017)]%
        {VapnikI17}
\bibfield{author}{\bibinfo{person}{Vladimir Vapnik} {and} \bibinfo{person}{Rauf
  Izmailov}.} \bibinfo{year}{2017}\natexlab{}.
\newblock \showarticletitle{Knowledge transfer in {SVM} and neural networks}.
\newblock \bibinfo{journal}{\emph{Ann. Math. Artif. Intell.}}
  \bibinfo{volume}{81}, \bibinfo{number}{1-2} (\bibinfo{year}{2017}),
  \bibinfo{pages}{3--19}.
\newblock


\bibitem[Vapnik and Vashist(2009)]%
        {VapnikV09}
\bibfield{author}{\bibinfo{person}{Vladimir Vapnik} {and}
  \bibinfo{person}{Akshay Vashist}.} \bibinfo{year}{2009}\natexlab{}.
\newblock \showarticletitle{A new learning paradigm: Learning using privileged
  information}.
\newblock \bibinfo{journal}{\emph{Neural Networks}} \bibinfo{volume}{22},
  \bibinfo{number}{5-6} (\bibinfo{year}{2009}), \bibinfo{pages}{544--557}.
\newblock


\bibitem[Williams and Seeger(2001)]%
        {Nystroem}
\bibfield{author}{\bibinfo{person}{Christopher K.~I. Williams} {and}
  \bibinfo{person}{Matthias Seeger}.} \bibinfo{year}{2001}\natexlab{}.
\newblock \showarticletitle{Using the Nystr\"{o}m Method to Speed Up Kernel
  Machines}.
\newblock In \bibinfo{booktitle}{\emph{Advances in Neural Information
  Processing Systems 13}}, \bibfield{editor}{\bibinfo{person}{T.~K. Leen},
  \bibinfo{person}{T.~G. Dietterich}, {and} \bibinfo{person}{V.~Tresp}} (Eds.).
  \bibinfo{publisher}{MIT Press}, \bibinfo{pages}{682--688}.
\newblock


\bibitem[Xu et~al\mbox{.}(2020)]%
        {rlNips}
\bibfield{author}{\bibinfo{person}{Zhiyuan Xu}, \bibinfo{person}{Kun Wu},
  \bibinfo{person}{Zhengping Che}, \bibinfo{person}{Jian Tang}, {and}
  \bibinfo{person}{Jieping Ye}.} \bibinfo{year}{2020}\natexlab{}.
\newblock \showarticletitle{Knowledge Transfer in Multi-Task Deep Reinforcement
  Learning for Continuous Control}. In \bibinfo{booktitle}{\emph{34th
  Conference on Neural Information Processing Systems (NeurIPS 2020)}}.
\newblock


\bibitem[Yim et~al\mbox{.}(2017)]%
        {yim2017gift}
\bibfield{author}{\bibinfo{person}{Junho Yim}, \bibinfo{person}{Donggyu Joo},
  \bibinfo{person}{Jihoon Bae}, {and} \bibinfo{person}{Junmo Kim}.}
  \bibinfo{year}{2017}\natexlab{}.
\newblock \showarticletitle{A gift from knowledge distillation: Fast
  optimization, network minimization and transfer learning}. In
  \bibinfo{booktitle}{\emph{Proceedings of the IEEE Conference on Computer
  Vision and Pattern Recognition}}. \bibinfo{pages}{4133--4141}.
\newblock


\bibitem[Yu et~al\mbox{.}(2017)]%
        {yu2017visual}
\bibfield{author}{\bibinfo{person}{Ruichi Yu}, \bibinfo{person}{Ang Li},
  \bibinfo{person}{Vlad~I Morariu}, {and} \bibinfo{person}{Larry~S Davis}.}
  \bibinfo{year}{2017}\natexlab{}.
\newblock \showarticletitle{Visual relationship detection with internal and
  external linguistic knowledge distillation}. In
  \bibinfo{booktitle}{\emph{Proceedings of the IEEE international conference on
  computer vision}}. \bibinfo{pages}{1974--1982}.
\newblock


\end{thebibliography}










\end{document}